\documentclass[twoside]{article}

\usepackage[preprint]{aistats2027}
\usepackage[round]{natbib}

\usepackage{amsmath}
\usepackage{amssymb}
\usepackage{amsthm}
\newtheorem{proposition}{Proposition}
\usepackage{booktabs}
\usepackage{multirow}
\usepackage{caption}
\usepackage{graphicx}
\usepackage{url}

\begin{document}

\twocolumn[

\aistatstitle{Verdict Instability of OOD Scores under Reference Resampling}

\aistatsauthor{ Donghoon Lee \And Shinjin Kang }

\aistatsaddress{ Hongik University \And  Hongik University} ]

\begin{abstract}
Post-hoc out-of-distribution detectors are fitted on a finite reference set, so every score they produce is an estimate. If we had chosen a different set, some verdicts would have moved. We measure that movement by resampling the reference set and recording the bootstrap standard deviation of the score, which we call verdict instability. It admits a closed form with no fitted parameters. The instability of a verdict is the within-class dispersion of the assigned class along the query's direction, divided by the square root of that class's reference count. That count is what separates verdict instability from the geometry of the score distribution, and it is identifiable only under class imbalance. Instability grows with the local dispersion. Far-OOD queries lie along the low-variance directions of an anisotropic embedding, so the distance-based scores in common use assign their highest values to the verdicts that are most
reproducible. Only estimators of local dispersion carry the sign a practitioner expects. We give a rule that predicts this sign for any score from a single label-free correlation, and abstention driven by a wrong-signed score turns out worse than abstention at random on every dataset we test.
\end{abstract}

\section{Introduction}
\label{sec:intro}

Post-hoc out-of-distribution (OOD) detectors are built from a finite reference set. A
Mahalanobis detector estimates class means and a covariance from it \citep{lee2018simple} and a
$k$NN detector stores it outright \citep{sun2022knn}. Even logit-based scores inherit the
reference set through the classifier head \citep{hendrycks2017baseline,liu2020energy}. The scores assigned to queries are therefore estimates. A different reference set would have changed the scores for some queries.

The question at a gate is not how far the query lies from the training data but how much its
verdict would move under a different draw of it. The two are routinely conflated, since OOD
scores are read as uncertainty estimates and used to abstain. Yet nobody has measured how much the finiteness of the reference set drives a score or whether proxy scores reflect it.

Fix a detector, resample the reference set by the bootstrap \citep{efron1994introduction} and
record the standard deviation of the score. We call this the query's verdict instability. It is
epistemic and needs no labels, so we can measure it where selective prediction \citep{geifman2017selective} cannot follow.

Our first result is that verdict instability admits a closed form. For a query at direction $u$
from the centroid of its assigned class $c$ with $n_c$ reference points,
\begin{equation}
\label{eq:main}
  \operatorname{std}\big[s(x)\big]
  \;\approx\;
  \sqrt{
    \frac{\sigma_t(u)^2}{n_c}
    \;+\;
    \lambda^2\, \mathrm{Var}\big[\mathrm{pen}(x)\big]
  } \, ,
\end{equation}
where $\sigma_t(u)^2 = u^{\top}\Sigma_c u$ projects the within-class scatter onto the query
direction and the second term is carried only by detectors with a global reference
\citep{ren2021simple}. Instability is the within-class dispersion of the assigned class along
the query's direction, divided by the square root of that class's reference count. There are no
free parameters. Eq.~\eqref{eq:main} tracks the bootstrap variance at $R^2 = 0.82$--$0.97$
across CIFAR-10, CIFAR-100, ImageNet-1k and DermaMNIST, with the class count running from $3$ to
$1000$.


\begin{figure*}[t]          
\centering
\includegraphics[width=0.75\textwidth]{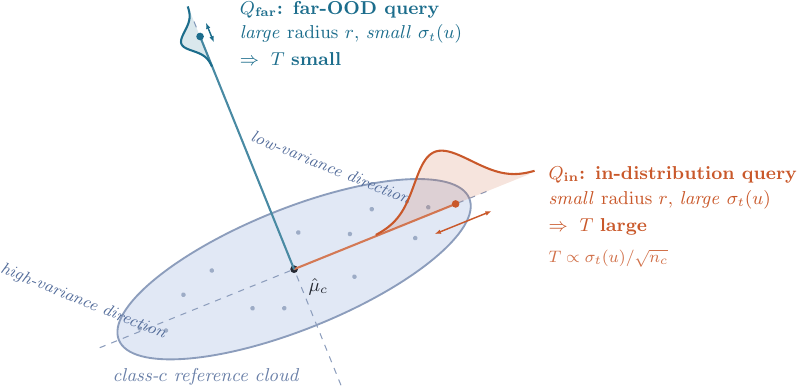}

\caption{\textbf{The geometry of the sign.} Verdict instability is the class cloud's dispersion
along the query direction, scaled by $n_c^{-1/2}$ (Eq.~\ref{eq:main}). $Q_{\text{in}}$ lies along a high-variance
eigen-direction and $Q_{\text{far}}$ along a low-variance one. The effect is the embedding's
anisotropy (\S\ref{sec:sign}), and removing the anisotropy removes the sign.}
\label{fig:inversion}

\end{figure*}

The count is what separates verdict instability from the geometry of the score distribution. No score in common use falls with it, and the recent geometric analysis of these detectors \citep{janiak2026geometry} does not either. We
therefore exploit the class imbalance of DermaMNIST \citep{yang2023medmnist, tschandl2018ham10000},
whose counts span a factor of $58.7$. Replacing $n_c$
by its mean collapses $R^2$ from $0.923$ to $0.276$.

Our second result follows from the first. Instability rises with $\sigma_t(u)$, so any score anti-correlated with the dispersion moves against the reliability of the verdict. Far-OOD queries sit along the low-variance directions of the class cloud, so the distance-based scores in common use assign their highest values to the queries whose verdicts are the most reproducible. Only the scores that estimate a dispersion directly carry the sign a practitioner expects. The sign is a property of the embedding rather than the theory, and it can be switched off by destroying the anisotropy of our self-supervised backbone \citep{caron2021emerging}.

The result is a rule that takes just one correlation to check. Its predictor is the model itself, and
$\operatorname{sign}[\rho(M, T)] = \operatorname{sign}[\rho(M, \widehat{T})]$ for any score $M$,
where the right-hand side needs no bootstrap and no labels. The count is essential here too, because under heavy imbalance a dispersion-only predictor reads the sign of several scores backwards where the full model does not. Abstaining on the queries a score flags as uncertain is then worse than random abstention, and this happens exactly when the score has the wrong sign. The wrong sign lands on a different score in each dataset, but the wrong-signed score fails in every case.

Our contributions are as follows.
\begin{itemize}
  \item We identify the bootstrap variance of an OOD score under reference-set resampling as the detector's epistemic uncertainty, and give a closed form with no fitted parameters
(Eq.~\ref{eq:main}).
  \item We validate the $1/\sqrt{n_c}$ dependence across classes under natural imbalance
    ($58.7\times$), which no other account of OOD scores depends on. We validate it again with the count assigned at random across ImageNet-1k, where it cannot stand for
the class.
  \item We separate distance from dispersion scores by sign. A label-free rule (Eq.~\ref{eq:rule}) predicts
the sign of any new score, including the stretch channel a geometric analysis tunes upward for
detection.
  \item Abstention driven by a wrong-signed score is worse than random. The rule calls the side of the
baseline correctly in all 15 cells over 10 seeds.
\end{itemize}
\section{Related Work}
\label{sec:related}

\paragraph{Post-hoc OOD scores.}
A large family of detectors attaches a scalar to a trained network without retraining. They read
the classifier head \citep{hendrycks2017baseline,liang2018enhancing}, the logits before
normalization \citep{liu2020energy,hendrycks2022scaling} or the features \citep{sun2022knn,ma2018lid,wang2022vim}, and surveys organize them into a common protocol
\citep{yang2022openood,yang2024generalized,fang2022learnable}. All are evaluated by how well they separate in-distribution inputs from the rest. We take the same 11 scores and
ask how each behaves against the uncertainty in its own detector.

\paragraph{The geometry of Mahalanobis detectors.}
The Mahalanobis family \citep{lee2018simple} has been refined at the score level by relative
Mahalanobis \citep{ren2021simple} and by Mahalanobis++ \citep{mueller2025mahalanobis}. A recent
line instead asks which properties of the feature space make the same quadratic detector succeed or
fail \citep{janiak2026geometry}, the closest work to ours. Its variance is taken over queries with
the reference set held fixed and does not depend on the reference count, while
ours is taken over reference sets with the query held fixed and is exactly $\Theta(n_c^{-1/2})$.
The two are built from the same geometry with reciprocal weights, and \S\ref{sec:related-geometry}
puts that work through our rule.

\paragraph{Uncertainty and its conflation with OOD.}
Predictive uncertainty splits into aleatoric and epistemic components \citep{kendall2017uncertainties}
and is estimated by MC dropout \citep{gal2016dropout}, deep ensembles
\citep{lakshminarayanan2017simple} or feature-space density \citep{mukhoti2023deep}. OOD scores are
routinely read as such estimates and used to abstain. That the conflation is unsafe is known in one
direction, since ReLU classifiers can be arbitrarily confident far from the data
\citep{ulmer2021know}. We argue the other. A score can detect well and still make a harmful
uncertainty proxy, and in this literature the uncertainty is always the model's and never the
detector's.

\paragraph{Finite samples and selective prediction.}
That a decision rule inherits uncertainty from the finite sample used to fit it is standard
statistics. The bootstrap \citep{efron1994introduction} is its canonical instrument and covariance
shrinkage \citep{ledoit2004well} is an admission that $\hat\Sigma$ is estimated, while conformal
prediction quantifies the same dependence for a calibration set
\citep{vovk2005algorithmic,angelopoulos2023conformal}. The corresponding statement for OOD scores
is what \S\ref{sec:theory} supplies. When a model may abstain, performance is summarized by the
risk--coverage curve and its area
\citep{elyaniv2010foundations,geifman2017selective,geifman2019bias}. That framework needs
ground-truth labels and is confined to the in-distribution region, whereas the failure mode we
identify lives where labels do not exist (\S\ref{sec:downstream}).
\section{Problem Formulation}
\label{sec:setup}

\subsection{Setup}

Let $f$ be a frozen feature extractor and $z = f(x) \in \mathbb{R}^{d}$ the embedding of $x$. A post-hoc detector is fitted on a labeled in-distribution reference set $\mathcal{R} = \{(x_i,
y_i)\}_{i=1}^{N}$, partitioned into class subsets $\mathcal{R}_c$ with
$n_c = \lvert \mathcal{R}_c \rvert$ and $\sum_c n_c = N$. Write $\hat\mu_c$ for the empirical class centroid, $\hat\mu_0$ for the global one and
$\hat\Sigma_c$ for the within-class scatter. Everything the detector knows about the in-distribution world it knows
through $\mathcal{R}$.

The detector assigns to a query $x$ a scalar
\begin{equation}
\label{eq:score}
  s(x; \mathcal{R})
  \;=\;
  \underbrace{\min_{c} \big\lVert z - \hat\mu_c \big\rVert}_{\text{class term}}
  \;+\;
  \lambda \cdot \underbrace{\big[\tau - \lVert z - \hat\mu_0 \rVert \big]_{+}}_{\text{global penalty}} ,
\end{equation}
where $[\,\cdot\,]_{+} = \max(0, \cdot)$ and $\tau$ is a low quantile of the global distance over
$\mathcal{R}$. The first term is the nearest-class-mean distance and the second is a hinge against
a global reference, in the spirit of the correction that relative Mahalanobis applies to the
class-conditional distance \citep{ren2021simple}. We analyze Eq.~(\ref{eq:score}) because it is
the simplest score carrying both ingredients that every member of the Mahalanobis family shares,
a class-conditional term estimated from $n_c$ points and a global term estimated from $N$.
Section~\ref{sec:sign} treats 11 standard scores as observables evaluated against its
instability rather than as objects to be re-derived. A verdict is obtained by thresholding at an
operating point $t$, so it is a statistic of $\mathcal{R}$ exactly as the score is.

\subsection{Verdict instability}

Fix the query and resample the reference set. Let $\mathcal{R}^{*}$ be a nonparametric bootstrap
replicate of $\mathcal{R}$ drawn class-wise so that the class counts $n_c$ are preserved
\citep{efron1994introduction}, and define
\begin{equation}
\label{eq:truth}
  T(x)
  \;\triangleq\;
  \operatorname{std}_{\mathcal{R}^{*}} \big[\, s(x; \mathcal{R}^{*}) \,\big] .
\end{equation}
We call $T(x)$ the verdict instability of $x$. It is the standard deviation of the score under a counterfactual draw of the reference data. It is
epistemic and caused by what was not collected. We will show
that it decays as $n_c^{-1/2}$. The standard deviation is taken over $B = 200$ replicates
throughout. The Monte-Carlo error this leaves in $T$ is common to every score compared against it
and is absorbed by the seed-level intervals of \S\ref{sec:worse}, since each seed redraws the
bootstrap.

Two properties make $T$ the object a practitioner actually needs. It converts directly into a
flip probability. If the score is approximately Gaussian under resampling, the chance that the
verdict at threshold $t$ reverses is
\begin{equation*}
  \Pr\big[\, \text{verdict flips} \,\big]
  \;\approx\;
  \Phi\!\left( - \frac{\lvert\, s(x; \mathcal{R}) - t \,\rvert}{T(x)} \right) ,
\end{equation*}
so instability is the natural denominator against which a margin should be read. A query far from the threshold in units of $T$ is safe and a query close to it is not, however far
it sits from the training data.

$T$ also requires no labels, neither for the query nor for the notion of correct. It is a property of the estimator rather than of the ground truth, so we can evaluate it in the
far-OOD region where the risk--coverage machinery of selective prediction
\citep{geifman2017selective, geifman2019bias} cannot follow.

$T$ is neither the probability that $x$ is OOD nor the model's predictive uncertainty. A confidently
wrong detector can be perfectly stable, so instability and correctness are orthogonal by
construction.

The remaining details of the evaluation protocol, the three query groups and the resolution and
class-count controls that keep the comparison honest, are given in Appendix~\ref{app:protocol}.
Correlations that treat $T$ as an observable to be tracked (\S\ref{sec:separation}) are computed
after centering both variables within query group, so that they report within-population behavior.
Section~\ref{sec:downstream} evaluates a global abstention policy instead, so its correlations are
computed over the pooled query set without centering. The two are reported separately and never
mixed.
\section{Verdict Instability as Projected Within-Class Dispersion}
\label{sec:theory}

The bootstrap quantity $T(x)$ of Eq.~(\ref{eq:truth}) costs $B$ refits, and a practitioner
holding a single reference set cannot obtain it at all. We predict it to within a few percent from that set alone. The score of Eq.~(\ref{eq:score}) has two
terms estimated from different samples, and each contributes an independent channel of variance.

\subsection{The class channel}
\label{sec:class-channel}

Fix a query $x$ with embedding $z$ and let $c$ be its assigned class. Write
$r = \lVert z - \hat\mu_c \rVert$ for its radius from the class centroid and
$u = (z - \hat\mu_c)/r$ for its direction. Under class-wise resampling the bootstrap centroid
obeys $\hat\mu_c^{*} = \hat\mu_c + n_c^{-1/2}\, \xi$ with
$\xi \rightsquigarrow \mathcal{N}(0, \Sigma_c)$. Substituting this into the class term of
Eq.~(\ref{eq:score}) and expanding the norm (Appendix~\ref{app:derivation}), the perturbation
enters at first order only through its component along the query direction. Retaining that term,
\begin{equation}
\label{eq:class-var}
  \operatorname{Var}\big[\, s_{\text{class}} \,\big]
  \;\approx\;
  \frac{\sigma_t(u)^2}{n_c} ,
\end{equation}
where $\sigma_t(u)^2 \triangleq u^{\top} \Sigma_c\, u$. We call $\sigma_t(u)$ the transverse
dispersion, the standard deviation of the class cloud projected onto the direction in which the
query happens to lie.

Three features of Eq.~(\ref{eq:class-var}) drive everything that follows. Only the assigned class matters at first order, because the other centroids move under resampling
and rarely attain the minimum. Proposition~\ref{prop:min} of Appendix~\ref{app:derivation} carries the case where one does. The dispersion is directional rather than global, since $\Sigma_c$ enters
only through the scalar $u^{\top}\Sigma_c u$. It is constant in an isotropic embedding and varies across the directions of an anisotropic one. And $\sigma_t(u)$ is a
plug-in functional of the reference set that needs no resampling and no labels.

\subsection{The penalty channel}
\label{sec:penalty-channel}

Write $D(x) = \lVert z - \hat\mu_0 \rVert$ for the global distance. It is estimated from all $N$
reference points, so its bootstrap fluctuation is $s_D \approx \sigma_w / \sqrt{N}$ with $\sigma_w$ the class-averaged dispersion along the global direction. This
channel is quiet at the source, but the hinge of Eq.~(\ref{eq:score}) rectifies it and
rectification does not preserve variance. Write $m = \tau - D(x)$ for the signed activation margin
and model $D^{*} \approx D + s_D \eta$. The penalty is then a rectified Gaussian whose variance is
available in closed form,
\begin{equation}
\label{eq:rect}
  \operatorname{Var}\big[\, s_{\text{pen}} \,\big]
  \;=\;
  \lambda^2 s_D^2 \cdot v\!\left( \frac{m}{s_D} \right) ,
\end{equation}
with $v$ the variance of a standard rectified Gaussian at margin $a$
(Appendix~\ref{app:derivation}). It is monotone in $a$ and vanishes as $a \to -\infty$, so a query outside the hinge contributes nothing and one at the elbow contributes most. Equation~(\ref{eq:rect}) is the whole of the penalty channel.

The penalty is therefore a tail phenomenon rather than a background term. On DermaMNIST most queries lie outside the hinge and contribute exactly zero, so the median penalty
share vanishes while the variance-weighted share is carried by the fraction at the elbow. The share also decays with the number of classes. It must, because the class channel scales as $1/n_c$
while the global channel scales as $1/(C\bar{n})$. Moving only the class-count threshold, the variance-weighted share falls monotonically with $C$ (Appendix~\ref{app:tables}).

\subsection{The two channels}
\label{sec:combined}

The two channels are driven by different samples. They are not strictly independent, since
$\hat\mu_0$ depends on class $c$ with weight $n_c/N$. We neglect the covariance and check the
cost empirically. Adding Eq.~(\ref{eq:class-var}) to Eq.~(\ref{eq:rect}) gives the explicit form of
Eq.~(\ref{eq:main}),
\begin{equation}
\label{eq:main-full}
  \widehat{T}(x)
  \;=\;
  \sqrt{
    \frac{\sigma_t(u)^2}{n_c}
    \;+\;
    \lambda^2 s_D^2 \cdot v\!\left( \frac{\tau - D(x)}{s_D} \right)
  } \; ,
\end{equation}
which has no fitted parameters. Here $\lambda$ and $\tau$ are hyperparameters of the detector
fixed before any instability is measured, and the remaining quantities are plug-in statistics of
$\mathcal{R}$. Nothing is tuned against $T$. We report a shape agreement and a scale agreement
because they are different claims (Table~\ref{tab:fit}).

\begin{table}[t]
\small
\centering
\caption{Equation~(\ref{eq:main-full}) against the bootstrap ground truth, with nothing fitted.
Here $R^2 = \mathrm{corr}(T,\widehat{T})^2$ measures shape agreement and the median of
$T/\widehat{T}$ measures scale agreement. Centering within query group costs at most 4 points of
$R^2$. The ImageNet row follows the protocol of Appendix~\ref{app:protocol} without change.}
\label{tab:fit}
\begin{tabular}{lccrc}
\toprule
Dataset & $C$ & $n_c$ & $R^2$ & $T/\widehat{T}$ \\
\midrule
CIFAR-10     & $5$    & $400$          & $0.918$ & $0.963$ \\
CIFAR-100    & $50$   & $400$          & $0.820$ & $0.953$ \\
ImageNet-1k  & $1000$ & $400$          & $0.910$ & $0.976$ \\
DermaMNIST   & $3$    & $769$--$4693$  & $\mathbf{0.974}$ & $1.005$ \\
DermaMNIST   & $7$    & $80$--$4693$   & $0.923$ & $0.976$ \\
\bottomrule
\end{tabular}
\end{table}

\noindent
The fit does not degrade as $C$ falls from $50$ to $3$ even though the composition of the variance
changes completely, and it sharpens as $C$ rises to $1000$. The channels trade off and the sum
tracks $T$ throughout. The scale is right without being fitted, since every median $T/\widehat{T}$
in Table~\ref{tab:fit} sits within a few percent of unity.

\subsection{The reference count}
\label{sec:scaling}

Equation~(\ref{eq:main-full}) carries one claim that separates it from every other analysis of OOD
scores. The class channel is a dispersion divided by a count, and the count is what
no score in common use falls with. Isolating it requires classes that differ in $n_c$ within a
single fit, sharing an embedding and a detector and a query set. The standard protocol fixes every $n_c$ and cannot supply this, and sweeping the reference-set size
moves every class at once. Natural imbalance can.

DermaMNIST \citep{yang2023medmnist, tschandl2018ham10000} supplies exactly this, with 7 lesion
classes at clinical frequencies and reference counts running from $80$ to $4693$. We do not subsample the majority classes down to the minority level. That would destroy the only
variation that identifies the count, and in a clinical reference set the common lesion is common.
We instead use the class-count threshold of \S\ref{sec:protocol} as a knob, which admits
progressively smaller classes and widens the imbalance from $6.1\times$ at $C = 3$ to
$58.7\times$ at $C = 7$.

The test is a single substitution. Replace the count in Eq.~(\ref{eq:main-full}) by its mean
$\bar{n} = N/C$ and leave every other quantity exactly as it was. The two models see the same
directions and the same dispersions and the same penalty. They differ only in whether a class
with $4693$ reference points is treated differently from one with $80$.

The gain grows with the imbalance, as it must if the count is doing real work. Replacing $n_c$ by
$\bar{n}$ drops the $R^2$ from $0.974$ to $0.756$ at $6.1\times$ and from $0.923$ to $0.276$ at
$58.7\times$, so at the widest imbalance discarding the count destroys $70\%$ of the explanatory power (Appendix~\ref{app:count}). A clinical frequency ties the count to the class,
so we seek imbalance a second way. Assigning the count at random across the 1000 classes of
ImageNet-1k makes the two independent by construction. The gain survives permuting which class receives which count (Appendix~\ref{app:nccontrol}). Once the reference set is
unbalanced, the count carries most of the content.

The exponent itself is not fitted but read off the class-conditional means. If
Eq.~(\ref{eq:class-var}) is right then
$\gamma_c \triangleq \overline{T}_c \sqrt{n_c} / \overline{\sigma}_{t,c}$ equals $1$ for every
class, the averages taken over queries assigned to $c$. This is a sharp test with one scalar per class and nothing tuned. Carrying the competing centroid in
closed form, the ratio holds within $0.08$ of unity from $80$ to $4693$ with no free parameter and
no channel dropped (Appendix~\ref{app:derivation}).

The dispersion form is what makes this test available at all. Our embeddings have $d = 768$ and
the smallest classes have $n_c < d$, so any whitened quantity would invert a singular scatter.
Equation~(\ref{eq:class-var}) never inverts $\hat\Sigma_c$. It evaluates the single quadratic
form $u^{\top}\hat\Sigma_c u$ whose coefficient of variation is $O(n_c^{-1/2})$ independently of
$d$, and the fit is undamaged at $d/n = 9.6$. This is a structural advantage of the dispersion
form over the whitened form to which we return in \S\ref{sec:related-geometry}.
\section{The Sign of a Score}
\label{sec:sign}

Verdict instability grows with $\sigma_t(u)$ and shrinks with $n_c$. Any score moving against that combination is anti-correlated with the reliability of its verdict, and a practitioner abstaining on it abstains on exactly the wrong queries. The standard scores are such quantities.

\begin{figure}[t]
\centering
\includegraphics[width=\columnwidth]{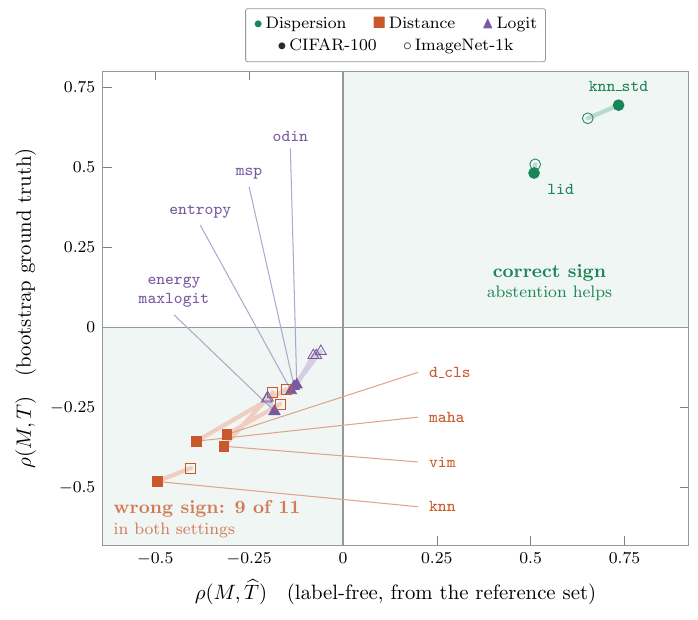}
\caption{\textbf{9 of 11 post-hoc scores carry the wrong sign, at both class counts.} Correlations
are centered within query group and a band joins the two settings of each score. Raising the class
count from $50$ to $1000$ moves none across a quadrant. All 22 points share the sign of both axes
(shaded), which \S\ref{sec:rule} turns into a rule.}
\label{fig:eleven}
\end{figure}

\subsection{The 11 scores against instability}
\label{sec:separation}

We evaluate 11 post-hoc scores as observables. For each score $M$ we report Spearman correlations with $\widehat{T}$ and with the bootstrap
instability $T$ of Eq.~(\ref{eq:truth}). Both are computed after centering within query group (\S\ref{sec:protocol}),
so that a correlation reports within-population behavior rather than the trivial fact that far-OOD queries differ from in-distribution ones on every axis.

\noindent
Figure~\ref{fig:eleven} is CIFAR-100 and ImageNet-1k, with the exact correlations in
Appendix~\ref{app:eleven}. The pattern is not a scatter of signs but a partition by what the score measures. Every score that estimates a distance is negatively correlated with instability, whether
it reads a class mean or the $k$-th neighbor or the whitened metric or the residual outside the
principal subspace. Only the two scores that estimate a local dispersion carry the positive sign,
and the logit family inherits the distance sign because a logit is a soft minimum over class
distances. That leaves 9 of 11 scores anti-correlated with the reliability of the verdict they
produce.

\begin{table}[t]
\small
\centering
\caption{Far queries lie along low-variance directions. The allocation effective dimension $1/\sum_i p_i^2$ rises from in- to far-OOD in every setting and
$\sigma_t$ falls with it. The in-distribution group carries the reversal of \S\ref{sec:why} on
DermaMNIST alone.}
\label{tab:groups}
\begin{tabular}{lrr}
\toprule
Query group & $\rho(r, \sigma_t)$ & Alloc.\ eff.\ dim. \\
\midrule
\multicolumn{3}{l}{\textit{CIFAR-100} ($C{=}50$)} \\
in & $-0.200$ & $36.5$ \\
near & $-0.269$ & $54.2$ \\
SVHN & $-0.138$ & $50.3$ \\
DTD & $-0.265$ & $76.2$ \\
LSUN & $-0.020$ & $52.5$ \\
iSUN & $-0.137$ & $55.4$ \\
Places365 & $-0.184$ & $56.3$ \\
CIFAR-10 & $-0.332$ & $56.1$ \\
\textit{pooled} & $\mathbf{-0.306}$ & --- \\
\midrule
\multicolumn{3}{l}{\textit{ImageNet-1k} ($C{=}1000$)} \\
in & $-0.210$ & $35.1$ \\
SSB-hard & $-0.065$ & $40.2$ \\
NINCO & $-0.105$ & $53.0$ \\
iNaturalist & $-0.225$ & $47.1$ \\
Textures & $-0.254$ & $67.8$ \\
OpenImage-O & $-0.147$ & $58.0$ \\
\textit{pooled} & $\mathbf{-0.223}$ & --- \\
\midrule
\multicolumn{3}{l}{\textit{DermaMNIST} ($C{=}3$)} \\
in & $+0.15$ & $\mathbf{8.5}$ \\
near & $+0.12$ & $12.9$ \\
SVHN & $-0.24$ & $34.7$ \\
DTD & $-0.61$ & $64.9$ \\
CIFAR-10 & $-0.47$ & $\mathbf{75.5}$ \\
\textit{pooled} & $\mathbf{-0.83}$ & --- \\
\bottomrule
\end{tabular}
\end{table}

\subsection{Why distance carries the negative sign}
\label{sec:why}

The sign is inherited from the embedding through one correlation. For a query far from the class
cloud the bootstrap fluctuation of the distance collapses onto one dimension of spread
$\sigma_t(u)$. The sign of any distance-like score is therefore the sign of $\rho(r, \sigma_t)$,
which asks whether far queries lie along high- or low-variance directions. Table~\ref{tab:groups}
shows they lie along the low-variance ones, both regimes in a single dataset.

\noindent
The second column of Table~\ref{tab:groups} is the mechanism, and it is worth stating in the
eigenbasis in which the geometry of these detectors has been analyzed
\citep{janiak2026geometry}. Write $\Sigma_c = E \Lambda E^{\top}$ and let
$p_i(z) = (e_i^{\top} u)^2$ be the allocation of the query's deviation
onto the $i$-th eigendirection, so that $\sum_i p_i = 1$. Then
\begin{equation}
\label{eq:alloc}
  \sigma_t(u)^2 \;=\; \sum_{i} p_i(z)\, \lambda_i ,
\end{equation}
the allocation-weighted mean of the spectrum. An in-distribution deviation is concentrated on the
few high-variance directions the embedding gave the class. A far-OOD deviation aligns with nothing in particular and spreads across a steeply
decaying spectrum, so its allocation is pulled toward the small eigenvalues. The allocation
effective dimension of Table~\ref{tab:groups} rises by nearly an order of magnitude from in- to
far-OOD data and $\sigma_t$ falls with
it. That is the whole of $\rho(r, \sigma_t) < 0$.

The in-distribution group is where the sign can go either way, and the second column of
Table~\ref{tab:groups} says which way. Its allocation effective dimension is $8.5$ on DermaMNIST
against the $36.5$ and $35.1$ the other two give, so a query that never leaves its class gains
radius only by running further along a high-variance direction and the correlation turns positive
there (Appendix~\ref{app:imagenet}). The in-group therefore reads the embedding rather than the
theory, and Eq.~(\ref{eq:main}) holds whatever the sign of $\rho(r, \sigma_t)$ turns out to be.
Section~\ref{sec:rule} shows it can be switched off.

\subsection{A falsification, and the rule}
\label{sec:rule}

If the account above is right, removing the anisotropy should erase the sign rather than weaken it. Whitening the embedding would equalize the eigenvalues, turning the transverse dispersion into a constant independent of the query direction and collapsing the correlations of the distance family with $\sigma_t$. Whitening changes the metric as well, so we isolate the mechanism with a weaker intervention. It is sufficient to fix where the deviation lies in the leading principal directions. This drives the pooled $\rho(r, \sigma_t)$ to zero in all three settings (Appendix~\ref{app:falsification}). The sign is not an artifact of our estimator. It is the anisotropy, and it can be switched off. The rule below is therefore conditional on the embedding. In an isotropic feature space there is no sign to predict, because no direction matters. The rule survives a change of training objective to CLIP in 63 of 66 cells and a change of
architecture family to ResNet-50 in 65. Under DINOv3, a more recent objective at identical input
geometry, it survives in all 66 (Appendix~\ref{app:backbone}).

What survives is a rule that costs no labels to check. The predictor is the model itself. Verdict
instability is estimated by $\widehat{T}$ of Eq.~(\ref{eq:main-full}), a plug-in functional of the
reference set. The claim is that its sign against a score matches that of the true instability,
\begin{equation}
\label{eq:rule}
  \operatorname{sign}\big[\, \rho(M, T) \,\big]
  \;=\;
  \operatorname{sign}\big[\, \rho(M, \widehat{T}) \,\big] .
\end{equation}
The predictor on the right-hand side needs no bootstrap and no labels, since $\widehat{T}$ is computed from the reference set alone. A practitioner can evaluate the rule for any candidate score
before deploying it. The rule is the model of \S\ref{sec:theory} used as a predictor, and it inherits the model's lack of
free parameters.

The count is the part of the model this criterion identifies. Weakening the predictor to
$\sigma_t^2 / n_c^{\alpha} + \lambda^2 \operatorname{Var}[\mathrm{pen}]$ and counting sign
agreements over 6 settings and 11 scores, dropping the count at $\alpha = 0$ costs 10 agreements out
of 66 that no penalty weight recovers, while the model's own values reach a plateau at the maximum
(Appendix~\ref{app:ablation}). The grid identifies the count, and Table~\ref{tab:fit} is where the
exponent and the penalty weight are read. The
count is essential because under heavy imbalance a logit reflects the class prior and so couples to
the $n_c$ in the denominator of instability, which a dispersion-only predictor cannot see.

\subsection{The rule predicts a quantity it was not built from}
\label{sec:related-geometry}

Equation~(\ref{eq:rule}) is a claim about any score, so we test it on one we did not design. Geometric analyses of Mahalanobis
detectors factor the score into a size channel and a whitened stretch channel
\citep{janiak2026geometry},
\begin{equation}
\label{eq:stretch}
  W(z) \;=\; u^{\top} \Sigma^{-1} u
  \;=\; \sum_{i} \frac{p_i(z)}{\lambda_i} ,
\end{equation}
and shape representations to control its variability for detection. Eq.~(\ref{eq:stretch}) inverts the spectrum that $u^{\top}\Sigma u$ averages over
the same allocation. Cauchy--Schwarz then gives
$W(z)\, u^{\top}\Sigma u \ge 1$, so stretch and dispersion are reciprocal proxies. The inequality is violated in none of the $14221$ queries the three settings hold, where $\rho(\log
W, -\log u^{\top}\Sigma u)$ runs from $0.83$ to $0.93$.

Equation~(\ref{eq:rule}) therefore predicts $W$ before we measure anything. Deferring by
stretch should be worse than deferring at random. It costs $2.44\%$ on CIFAR-100, $0.87\%$ on ImageNet-1k and $12.75\%$ on DermaMNIST against a coin
flip (Appendix~\ref{app:stretch}). The ImageNet margin is the narrowest because
$\rho(W,\widehat{T})$ sits nearest zero there. The size channel and the Mahalanobis score behave the same way. That work targets detection performance and achieves it. Our statement is about what the target
omits. In this embedding the quantity one maximizes to separate in- from out-of-distribution inputs is the
reciprocal of the quantity that governs whether the verdict survives a different reference set.
\section{The Cost of the Wrong Sign}
\label{sec:downstream}

Section~\ref{sec:sign} established a sign and a rule that predicts it without labels. This section
measures what the sign costs a practitioner who ignores it. Abstention driven by a wrong-signed
score is worse than abstention driven by nothing at all, and which scores are wrong-signed is what
Eq.~(\ref{eq:rule}) says in advance.

\subsection{Instability--coverage}
\label{sec:ic-curve}

Fix a score $M$ used to abstain. At coverage $\kappa$ we retain the fraction $\kappa$ of queries on
which $M$ is smallest and record the mean verdict instability of the retained set. Sweeping $\kappa$
from $1$ to $0.5$ traces an instability--coverage curve whose area (AURC) summarizes how much stability the abstention buys, in analogy with the risk--coverage curves of selective prediction
\citep{geifman2017selective, geifman2019bias}. Random abstention is the baseline and its curve is flat, since removing a random subset leaves the
mean unchanged. A useful score bends the curve down and a harmful one bends it up. That flatness fixes the baseline at $\tfrac{1}{2}\overline{T}$, so a different draw of queries
rescales every row together. We therefore report the position of a score relative to the baseline as $\Delta_M = \mathrm{AURC}_M / \mathrm{AURC}_{\mathrm{rand}} - 1$, which the rescaling leaves fixed.

The curve needs no labels, which is the point. Risk--coverage is defined only where correctness is defined, so it cannot be evaluated in the
far-OOD region at all. Ours can, because the target is a property of the estimator rather than of
the truth.

\begin{table*}[t]
\small
\centering
\caption{Abstention against verdict instability on three datasets. $\Delta$ is the position of a
score relative to the random baseline in percent, so a positive value (bold) costs more than random.
Entries average 10 seeds and carry half the width of a $95\%$ interval over them. Correlations are
pooled without centering (\S\ref{sec:setup}).}
\label{tab:worse}
\resizebox{\textwidth}{!}{%
\begin{tabular}{l rrl c rrl c rrl}
\toprule
& \multicolumn{3}{c}{\textit{CIFAR-100} ($C{=}50$)} && \multicolumn{3}{c}{\textit{ImageNet-1k} ($C{=}1000$)} && \multicolumn{3}{c}{\textit{DermaMNIST} ($C{=}3$)} \\
\cmidrule(lr){2-4}\cmidrule(lr){6-8}\cmidrule(lr){10-12}
Score & $\rho(M,\widehat{T})$ & AURC & $\Delta$ (\%) && $\rho(M,\widehat{T})$ & AURC & $\Delta$ (\%) && $\rho(M,\widehat{T})$ & AURC & $\Delta$ (\%) \\
\midrule
\texttt{knn\_std} & $+0.714$ & 0.1201 & $-6.99${\scriptsize$\,\pm0.14$} && $+0.631$ & 0.1376 & $-8.82${\scriptsize$\,\pm0.17$} && $+0.747$ & 0.0500 & $-19.70${\scriptsize$\,\pm0.20$} \\
\texttt{lid} & $+0.541$ & 0.1220 & $-5.57${\scriptsize$\,\pm0.16$} && $+0.545$ & 0.1406 & $-6.81${\scriptsize$\,\pm0.13$} && $+0.714$ & 0.0500 & $-19.68${\scriptsize$\,\pm0.20$} \\
\texttt{energy} & $-0.108$ & 0.1317 & $\mathbf{+1.95}${\scriptsize$\,\pm0.20$} && $-0.220$ & 0.1554 & $\mathbf{+3.00}${\scriptsize$\,\pm0.17$} && $+0.530$ & 0.0533 & $-14.26${\scriptsize$\,\pm0.30$} \\
\texttt{msp} & $-0.193$ & 0.1321 & $\mathbf{+2.30}${\scriptsize$\,\pm0.11$} && $-0.134$ & 0.1537 & $\mathbf{+1.87}${\scriptsize$\,\pm0.13$} && $+0.320$ & 0.0580 & $-6.84${\scriptsize$\,\pm0.38$} \\
\texttt{maha} & $-0.534$ & 0.1342 & $\mathbf{+3.93}${\scriptsize$\,\pm0.11$} && $-0.206$ & 0.1557 & $\mathbf{+3.19}${\scriptsize$\,\pm0.14$} && $-0.696$ & 0.0704 & $\mathbf{+13.09}${\scriptsize$\,\pm0.18$} \\
\midrule
\textit{random} & \textit{$+0.000$} & \textit{0.1291} & \textit{---} && \textit{$+0.000$} & \textit{0.1509} & \textit{---} && \textit{$+0.000$} & \textit{0.0622} & \textit{---} \\
\bottomrule
\end{tabular}}
\end{table*}

\subsection{Worse than random}
\label{sec:worse}

\noindent
Deferring the queries an energy detector
calls uncertain leaves behind a population whose verdicts are less reproducible than the one it
started with, and the retained instability rises monotonically as coverage falls while random
abstention holds flat.

The sign of $\rho(M,\widehat{T})$ says which side of the baseline a score lands on, and it is right
for all 5 scores on all three datasets. The interval on $\Delta$ excludes zero in all 15 cells and
its sign agrees in all 10 seeds. The narrowest margin is MSP on ImageNet-1k at $+1.87\%$ with an interval of $[+1.73, +2.00]$. The level $\overline{T}$ moves between runs and the contrast does not
(Appendix~\ref{app:seeds}).

The magnitudes order the extremes as well. The score most strongly aligned with the class channel
buys the most stability and the score most strongly anti-aligned costs the most. Ranking by $\rho(M,\widehat{T})$ reproduces the ranking by $\Delta$ on all three datasets wherever
the two scores differ by more than the interval on $\Delta$. The pairs it leaves unresolved sit
within $0.35$ percentage points of each other.

Sweeping the reference count leaves every score on the same side of the baseline, while the mean
instability falls as $n_c^{-1/2}$ (Appendix~\ref{app:sweep}). Collecting more reference data shrinks the
instability without repairing the sign.

\subsection{What moves the failing score}
\label{sec:derma-flip}

Table~\ref{tab:worse} might be read as a verdict on energy. The failure travels with the sign rather
than with the detector, and what moves the sign is the reference set.

\noindent
Raising the class count does not move it. Between CIFAR-100 and ImageNet-1k the held-in classes go from $50$ to $1000$ and the images from $32$ pixels to natural resolution, and every score keeps its
side of the baseline. Mahalanobis carries the largest loss in both at $+3.93\%$ and $+3.19\%$, and the two dispersion
scores buy the most stability in both. Figure~\ref{fig:eleven} shows the same invariance upstream.

Imbalance moves it. The coupling of \S\ref{sec:rule} reaches only the scores that read the class
prior, so on DermaMNIST energy and MSP cross the baseline and buy stability where they cost it on
the other two. Mahalanobis never touches the count, so it keeps the failure it carries everywhere
and the largest one here at $+13.09\%$. The set of failing scores changed while the law did not.
What decides which detector fails is therefore the same quantity that
\S\ref{sec:scaling} identifies, read downstream.

The scores that fail here are optimized against a different target, and
the two targets turn out close to antipodal. Evaluated against misclassification of the
in-distribution classifier in the conventional way, the same scores invert their order almost exactly, so each is beaten by a coin flip in the column it
was not designed for
(Appendix~\ref{app:orthogonal}). At least three separate quantities travel under the one name of
uncertainty, how likely the classifier is to be wrong and how unfamiliar the input is and how
reproducible the verdict is.
\section{Limitations}
\label{sec:limitations}

Three boundaries fix what the result covers. The estimand is the instability of
Eq.~(\ref{eq:score}), and \S\ref{sec:sign} tracks the 11 standard scores against it rather than resampling each. The class channel is visible only in an unbalanced reference set, and we
read it from counts of $30$ and above (Appendix~\ref{app:nccontrol}). The sign belongs to the embedding rather than to the theory. It switches off under the intervention of \S\ref{sec:rule}, it weakens as the far-OOD sources come closer and it turns over where the in-distribution allocation
is narrow enough (Appendix~\ref{app:imagenet}). Instability is orthogonal to correctness. A
reproducible verdict is not a right one.

\section{Conclusion}
\label{sec:conclusion}

An out-of-distribution score is an estimate fitted on a reference set that could have been drawn
differently. We asked how much a verdict would move if it had been. The answer has a closed form
with no fitted parameters, the within-class dispersion of the assigned class along the query's
direction divided by the square root of that class's reference count. The count is what separates
this quantity from the geometry of the score distribution, and it is identifiable only under class imbalance. Its consequence is a sign. In an anisotropic embedding the far queries lie along the
low-dispersion directions, so the distance-based scores are anti-correlated with the reliability of
the verdicts they produce. Only the dispersion scores carry the sign a practitioner expects. The
sign costs one label-free correlation to check, and deferring on a wrong-signed score is worse than
deferring at random.

At least three quantities travel under the one name of uncertainty, and on real representations they are oppositely ordered rather than merely distinct. One of them is verdict instability. It is measurable and it is predictable, and it remains
unmeasured. We argue that it
belongs in the budget.




\bibliography{references}

\clearpage
\appendix
\thispagestyle{empty}

\onecolumn

\section{Evaluation Protocol}
\label{app:protocol}
\label{sec:protocol}

\paragraph{Datasets.}
The in-distribution sets are CIFAR-100 and CIFAR-10 \citep{krizhevsky2009learning} and DermaMNIST
\citep{yang2023medmnist, tschandl2018ham10000}. In the CIFAR settings the far-OOD sources are SVHN
\citep{netzer2011svhn}, DTD \citep{cimpoi2014describing}, LSUN \citep{yu15lsun}, iSUN
\citep{xu15arXiv}, Places365 \citep{zhou2017places} and the other CIFAR set. DermaMNIST draws its
far-OOD from SVHN, DTD and CIFAR-10 alone. Six settings recur throughout. Five are CIFAR-100 at $C = 50$,
CIFAR-10 at $C = 5$ and DermaMNIST at $C = 3, 5, 7$. The sixth is ImageNet-1k
\citep{deng2009imagenet} under the OpenOOD v1.5 benchmark \citep{yang2022openood}, which supplies
the near sources SSB-hard and NINCO \citep{bitterwolf2023ninco} and the far sources iNaturalist,
Textures and OpenImage-O. Its reference set holds the same $400$ points per class as the CIFAR
settings, drawn from a pool of $500$, and every detector hyperparameter below is carried over
unchanged. ImageNet-1k has no held-out classes, so the OpenOOD near pair stands where the other
settings put a held-out split. Every source is public and used as released. DermaMNIST repackages HAM10000, a collection of
de-identified dermatoscopic images its curators released for research.

\paragraph{Backbone.}
The frozen feature extractor is a ViT-B/16 trained with DINO \citep{caron2021emerging}, taken as the
public \texttt{timm} checkpoint \texttt{vit\_base\_patch16\_224.dino}. It carries no classification
head and gives $d = 768$. Images are encoded at $224 \times 224$ under that checkpoint's default
evaluation transform, a resize to $248$ followed by a center crop and ImageNet normalization. The
three check encoders of Appendix~\ref{app:backbone} are CLIP
(\texttt{vit\_base\_patch16\_clip\_224.openai}), ResNet-50 (\texttt{resnet50.a1\_in1k}) at
$d = 2048$, and DINOv3 (\texttt{vit\_base\_patch16\_dinov3.lvd1689m}). All four share the resize
and crop geometry above. CLIP matches it natively. ResNet-50 and DINOv3 are overridden, the latter
by building the transformer at $224$ so that its position embeddings interpolate from the $256$ it
was trained at. The same crop of the same image therefore enters every encoder, while the
normalization constants remain each backbone's own. Nothing is fine-tuned.

\paragraph{Query groups.}
Queries are drawn from three disjoint groups. These are held-in test data, near-OOD (held-out
classes of the same dataset) and far-OOD (unrelated datasets). The groups differ enormously in both
$s$ and $T$, so a pooled correlation between any score and $T$ is dominated by between-group
structure. Correlations that track $T$ as an observable (\S\ref{sec:separation}) therefore center
both variables within each group. The abstention experiments of \S\ref{sec:downstream} rank the
pooled query set instead, because that is the policy a practitioner deploys. Their correlations
pool the same way and skip the centering. Each table states which convention it follows, and we
never compare one against the other.

\paragraph{Resolution control.}
Because our claims compare in-distribution and OOD populations inside one embedding, where
conclusions are known to be fragile to contamination \citep{bitterwolf2023ninco} and to
representation choice \citep{fort2021exploring}, any systematic difference in preprocessing between
them is a confound. When a $28 \times 28$ medical dataset is paired against natural-image OOD sets,
every OOD image is downsampled to $28 \times 28$ before feature extraction, so that upsampling
artifacts cannot be mistaken for distributional distance. We report the mean feature norm of each
group as a check.

\paragraph{Class filtering.}
Held-in classes are selected by a sample-count threshold and the classes that fall below it are
recycled as near-OOD rather than discarded. We do not subsample majority classes down to the
minority level, because imbalance is the deployed condition and is the only setting in which the
$n_c$ dependence of Eq.~(\ref{eq:truth}) is identifiable at all (\S\ref{sec:scaling}).

\paragraph{Detector hyperparameters.}
The detector of Eq.~(\ref{eq:score}) runs at $\lambda = 5$, with $\tau$ the $20$th percentile of the
global distance over $\mathcal{R}$. Both are fixed before any instability is measured. The 11 scores
of \S\ref{sec:separation} take their conventional settings: \texttt{knn} reads the $5$th neighbor,
\texttt{lid} the maximum-likelihood estimate over $20$, \texttt{vim} a principal subspace of
dimension $64$, \texttt{maha} a shrinkage of $0.3$ toward the scaled identity
\citep{ledoit2004well}, and \texttt{knn\_std} the standard deviation over a window of $0.7\,n_c$
neighbors inside the assigned class. The logit family reads a multinomial logistic probe fitted on
$\mathcal{R}$ (Appendix~\ref{app:odin}) at its default $\ell_2$ regularization, and \texttt{odin}
uses the published $T = 1000$ and $\varepsilon = 0.0014$. The probe is optimized to a gradient
tolerance of $10^{-6}$ rather than the $10^{-4}$ its solver defaults to. The default halts at an
objective a third above the optimum and reports no warning, which leaves the logit scores standing
wherever the optimizer stopped. Every number in this paper uses the converged fit. Reference
sets hold $400$ points per class in the CIFAR settings and the full training split on DermaMNIST,
and each query group contributes at most $800$ queries on CIFAR and $700$ on DermaMNIST. None of
these values is selected against $T$.

\paragraph{Seeds.}
Every number in \S\ref{sec:downstream} is a mean over 10 seeds. A seed redraws the reference
subsample, the query subsample, the $B$ bootstrap replicates and the random baseline. The intervals
we report are therefore end-to-end rather than conditional on a fixed draw. DermaMNIST keeps the
full training split as its reference set by design, because subsampling it would destroy the natural
imbalance the identification depends on. A seed there varies the query subsample and the bootstrap
alone. Query counts are $6400$ on CIFAR-100, $4800$ on ImageNet-1k and $3021$ on DermaMNIST.

\paragraph{Compute.}
Everything runs on one workstation with a single NVIDIA RTX 4090 of $24$ GB, $16$ cores and
$30$ GB of system memory available to the environment. Feature extraction is the only step that
loads the GPU and takes a few minutes per encoder over the $1.7 \times 10^{5}$ images the study
encodes. The exception is \texttt{odin}, which recomputes an embedding from raw pixels with an
input gradient and takes about 20 minutes per encoder. Once the features are cached the rest
runs on CPU, since a bootstrap replicate recomputes only the centroids and
Eq.~(\ref{eq:score}).

\section{Derivation of the Two Channels}
\label{app:derivation}

\paragraph{Assumptions.}
Everything below rests on eight statements, and we list them so that each can be checked
against the experiments. (A1) Class-wise resampling leaves the bootstrap centroids independent
across classes and each obeys the central limit behavior. (A2) Every $n_c$ is large enough for
that behavior, which \S\ref{sec:scaling} reads from counts of $30$ and above. (A3) The query does
not sit at a centroid, so the radius $r$ is positive. (A4) The quadratic term of
Eq.~(\ref{eq:expand}) is negligible at the scale $\epsilon$ defined below. (A5) Only the nearest
and the runner-up centroid compete for the minimum. (A6) The two channels are uncorrelated, which
\S\ref{sec:combined} measures rather than assumes. (A7) The global distance is approximately
Gaussian under resampling and $\tau$ is held fixed. (A8) The empirical scatter $\hat\Sigma_c$ stands for the population one.
The smallest classes have $n_c < d$, so that estimate is rank deficient, and \S\ref{sec:scaling}
is where the design avoids inverting it.

Assumption (A5) is the one the score itself can break, because a replicate may change which
centroid attains the minimum. The minimum is differentiable wherever the argmin is unique, and the
nonsmooth points are the ties. A nonparametric bootstrap can be inconsistent for a functional of
that kind \citep{dumbgen1993nondifferentiable}, the general treatment sets it among the
directionally differentiable ones \citep{fang2019directionally} and the standard remedy resamples
fewer than $n$ points \citep{bickel1997resampling}. We take neither remedy, because (A5) confines
the competition to two classes and Proposition~\ref{prop:min} carries that case in closed form.
The switch probability of the proposition measures how far the assignment moves, and
Table~\ref{tab:mincorr} puts it against the bootstrap.

\paragraph{The class channel.}
Fix a query $x$ with embedding $z$ and assigned class $c$. Write
\begin{equation*}
  r \;=\; \big\lVert z - \hat\mu_c \big\rVert ,
\end{equation*}
for the radius and $u = (z - \hat\mu_c)/r \in \mathbb{S}^{d-1}$ for the direction. Under
class-wise bootstrap resampling the centroid obeys the central limit behavior
\begin{equation*}
  \hat\mu_c^{*} \;=\; \hat\mu_c \;+\; \frac{1}{\sqrt{n_c}}\, \xi ,
  \qquad
  \xi \;\rightsquigarrow\; \mathcal{N}\!\big(0,\, \Sigma_c\big) .
\end{equation*}
Substituting into the class term of Eq.~(\ref{eq:score}) and expanding the norm,
\begin{equation}
\label{eq:expand}
  \big\lVert z - \hat\mu_c^{*} \big\rVert
  \;=\;
  r
  \;-\; \frac{1}{\sqrt{n_c}}\, \langle u, \xi \rangle
  \;+\; \frac{1}{2 r\, n_c}\Big( \lVert \xi \rVert^2 - \langle u, \xi\rangle^2 \Big)
  \;+\; O\!\big( n_c^{-3/2} \big) .
\end{equation}
The perturbation enters at first order only through its component along the query direction.
Retaining that term and using $\operatorname{Var}[\langle u, \xi\rangle] = u^{\top}\Sigma_c u$
gives Eq.~(\ref{eq:class-var}) of the main text. The dropped quadratic term is $O(r\epsilon^2)$
with $\epsilon = \sqrt{\operatorname{tr}\Sigma_c / n_c}\,/\,r$, which stays under $0.4\%$ everywhere.

\paragraph{The competing centroid.}
Write $(1)$ and $(2)$ for the nearest and the runner-up class of a query, $s_i^2 =
\sigma_{t,(i)}^2 / n_{(i)}$ for their first-order variances, $\theta^2 = s_1^2 + s_2^2$ and
$\alpha = (d_{(2)} - d_{(1)})/\theta$ for the gap in units of that fluctuation.

\begin{proposition}
\label{prop:min}
Under (A1)--(A5) the class channel is
\begin{equation}
\label{eq:mincorr}
  \operatorname{Var}\big[\, \min_c d_c^{*} \,\big]
  \;=\;
  s_1^2\, \Phi(\alpha) \;+\; s_2^2\, \overline{\Phi}(\alpha) \;+\; \theta^2 g(\alpha) ,
\end{equation}
where $\overline{\Phi} = 1 - \Phi$ and
\begin{equation*}
  g(\alpha) \;=\; \alpha^2 \overline{\Phi}(\alpha) \;-\; \alpha\, \phi(\alpha)
  \;-\; \big[\, \alpha \overline{\Phi}(\alpha) - \phi(\alpha) \,\big]^2 .
\end{equation*}
A replicate changes the argmin with probability $\overline{\Phi}(\alpha)$.
\end{proposition}

\begin{proof}
Retaining the first-order term of Eq.~(\ref{eq:expand}) makes each $d_{(i)}^{*}$ Gaussian with
mean $d_{(i)}$ and variance $s_i^2$, and (A1) makes the two independent. Abbreviate $D_i = d_{(i)}^{*}$ and write
$\min(D_1, D_2) = -\max(-D_1, -D_2)$. Read the first two moments of the maximum of two
Gaussians off \citet{clark1961greatest}, which for independent arguments gives
$\theta^2 = s_1^2 + s_2^2$ and $\alpha = (d_{(2)} - d_{(1)})/\theta$. Subtracting the square of the
first moment from the second leaves Eq.~(\ref{eq:mincorr}). The argmin changes exactly when
$D_2 < D_1$, whose probability is $\overline{\Phi}(\alpha)$.
\end{proof}

Two limits fix the reading. As $\alpha \to \infty$ the gap swallows the fluctuation and
Eq.~(\ref{eq:mincorr}) returns $s_1^2$, which is Eq.~(\ref{eq:class-var}) of the main text. At $\alpha = 0$ it returns $(s_1^2 + s_2^2)(1/2 - 1/2\pi)$, which falls below $s_1^2$ only while
$s_2 < 1.39\, s_1$. Competition therefore narrows the spread when the two classes fluctuate alike,
by $17.4\%$ at $s_1 = s_2$, and widens it once the runner-up is the noisier of the two.

That asymmetry is what moves both endpoints of Table~\ref{tab:gamma} and moves them in opposite
directions. A query assigned to a small class carries a large $s_1$ against a quieter runner-up, so
the minimum compresses its spread and $T$ falls below the uncorrected prediction. A query assigned
to the majority class carries the reverse and its spread widens, which is the inflation
Appendix~\ref{app:tables} reports. The exponent read off the sweep comes out below one half for the
same reason. Table~\ref{tab:mincorr} puts the
prediction against the bootstrap without holding the argmin anywhere.

\begin{table}[htbp]
\small
\centering
\caption{Proposition~\ref{prop:min} against the bootstrap. The switch rate is the share of
replicates that change the argmin, predicted by $\overline{\Phi}(\alpha)$ and measured directly,
with the correlation taken over queries. The last four columns are the fit of
Eq.~(\ref{eq:main-full}) before and after the correction, at $B = 200$ and with nothing fitted.}
\label{tab:mincorr}
\begin{tabular}{lcccrrrr}
\toprule
& \multicolumn{3}{c}{Switch rate} && \multicolumn{1}{c}{$R^2$} && \multicolumn{1}{c}{med $T/\widehat{T}$} \\
\cmidrule(lr){2-4}\cmidrule(lr){6-6}\cmidrule(lr){8-8}
Setting & obs & pred & $\rho$ && plain\;$\to$\;corr && plain\;$\to$\;corr \\
\midrule
CIFAR-100    & $8.1\%$  & $7.3\%$ & $+0.979$ && $0.820 \to \mathbf{0.922}$ && $0.953 \to 0.971$ \\
CIFAR-10     & $5.2\%$  & $4.8\%$ & $+0.990$ && $0.918 \to \mathbf{0.934}$ && $0.963 \to 0.979$ \\
ImageNet-1k  & $7.0\%$  & $6.4\%$ & $+0.981$ && $0.910 \to \mathbf{0.965}$ && $0.976 \to 0.991$ \\
Derma $C{=}3$ & $4.5\%$  & $4.5\%$ & $+0.992$ && $0.974 \to \mathbf{0.982}$ && $1.005 \to 1.016$ \\
Derma $C{=}5$ & $6.1\%$  & $5.8\%$ & $+0.987$ && $0.973 \to \mathbf{0.985}$ && $0.967 \to 0.983$ \\
Derma $C{=}7$ & $9.9\%$  & $9.0\%$ & $+0.968$ && $0.923 \to \mathbf{0.984}$ && $0.976 \to 0.980$ \\
\bottomrule
\end{tabular}
\end{table}

\noindent
The switch rate is predicted to within about one point in every setting and the query-level
correlation runs from $0.968$ to $0.992$, so the competition of Table~\ref{tab:gamma} is a quantity
the model computes rather than a name we give the residual. The correction helps most where the
uncorrected fit was weakest. Assumption (A5) is tightest where it is most at risk, since
ImageNet-1k offers $999$ competitors for the minimum and the prediction holds there as well as it
does at $C = 3$.

Read class by class, the correction removes the drift at both ends without holding the argmin
anywhere. On DermaMNIST at $C = 7$ the ratio $\overline{T}_c / \overline{\widehat{T}}_c$ runs
$0.797$, $0.897$, $0.980$, $0.973$, $1.005$, $0.993$, $1.034$ over the seven classes before the
correction and $0.931$, $0.969$, $1.012$, $0.980$, $0.975$, $0.973$, $0.902$ after it, so a spread
of $0.237$ falls to $0.110$.

\paragraph{The penalty channel.}
Let $m = \tau - D(x)$ be the signed activation margin and $s_D$ the bootstrap standard deviation
of the global distance.

\begin{proposition}
\label{prop:pen}
Under (A7) the penalty channel is $\lambda^2 s_D^2 \, v(m/s_D)$ with
\begin{equation*}
  v(a) \;=\; (a^2 + 1)\,\Phi(a) \;+\; a\,\phi(a) \;-\; \big[\, a\,\Phi(a) + \phi(a) \,\big]^2 ,
\end{equation*}
where $\phi$ and $\Phi$ are the standard normal density and distribution function.
\end{proposition}

\begin{proof}
Write $D^{*} = D + s_D \eta$ with $\eta \sim \mathcal{N}(0,1)$ by (A7), so the penalty is
$\lambda s_D [\, a - \eta \,]_{+}$ at $a = m/s_D$. Integrating the rectified variable against the
standard normal density gives $\mathbb{E}[(a-\eta)_{+}] = a\Phi(a) + \phi(a)$ and
$\mathbb{E}[(a-\eta)_{+}^2] = (a^2+1)\Phi(a) + a\phi(a)$. Subtracting the square of the first from
the second leaves $v(a)$, and the factor $\lambda^2 s_D^2$ restores the scale.
\end{proof}

\noindent
The function $v$ is monotone in $a$ and vanishes as $a \to -\infty$.

\section{Additional Tables}
\label{app:tables}

\paragraph{Penalty share and the imbalance knob.}
With the dataset and the embedding and the resolution fixed and only the class-count threshold of
\S\ref{sec:protocol} moving, the variance-weighted share of $\operatorname{Var}[s]$ carried by the
penalty channel decays with $C$ as Eq.~(\ref{eq:main-full}) requires. The median share is $0$ at every $C$,
since $94\%$ of queries never activate the hinge. The same threshold serves as an imbalance knob,
because classes below it are recycled as near-OOD rather than discarded, so lowering it admits
progressively smaller classes and widens the imbalance without any subsampling.

\begin{table}[htbp]
\small
\centering
\caption{The class-count threshold governs two quantities at once. As it admits smaller classes the
imbalance widens without subsampling, and the variance-weighted penalty share decays with $C$ as the
$1/(C\bar{n})$ scaling of the global channel requires. The median share is $0$ throughout. The first
column is ImageNet-1k, which carries no imbalance to widen and sits at the far end of the same $C$
trend.}
\label{tab:knob}
\begin{tabular}{lcccc}
\toprule
\texttt{min\_n}                 & ---        & $400$       & $200$        & $80$ \\
Held-in classes $C$             & $1000$     & $3$         & $5$          & $7$ \\
Imbalance $\max n_c / \min n_c$ & $1\times$  & $6.1\times$ & $20.6\times$ & $\mathbf{58.7\times}$ \\
\midrule
Variance-weighted penalty share  & $0.6\%$    & $34.9\%$    & $21.9\%$     & $9.8\%$ \\
Median per-query share           & $0\%$      & $0\%$       & $0\%$        & $0\%$ \\
\bottomrule
\end{tabular}
\end{table}

\noindent
Query groups are held fixed across the 3 settings in the main text, so that only the
reference partition moves. Letting the groups vary naturally with the threshold changes nothing
material. The fit is unchanged. It reads $R^2 = 0.968$ and $0.978$ and $0.912$ at $C = 3, 5, 7$ against the
$0.974$ and $0.973$ and $0.923$ obtained with groups held fixed.

\paragraph{Reading the exponent off the classes.}
The ratio $\gamma_c = \overline{T}_c \sqrt{n_c} / \overline{\sigma}_{t,c}$ equals $1$ if the class
channel scales as $n_c^{-1/2}$. Evaluated on all queries it returns $1.77$ in the majority class,
which is the penalty channel being misattributed to the scaling. The penalty-inactive subset
removes the contamination.

\begin{table}[htbp]
\small
\centering
\caption{The exponent read off the classes, on DermaMNIST. The ratio
$\gamma_c = \overline{T}_c \sqrt{n_c} / \overline{\sigma}_{t,c}$ equals $1$ if the class channel
scales as $n_c^{-1/2}$. Evaluated on all queries it returns $1.77$ in the majority class, which
is the penalty channel being misattributed to the scaling. The second row holds the argmin at its
observed assignment. A balanced setting gives every class the same $n_c$ and the ratio collapses to
one number, so Table~\ref{tab:ncgamma} reads the same quantity off ImageNet-1k with the count
assigned at random.}
\label{tab:gamma}
\begin{tabular}{lccccccc}
\toprule
$n_c$ & $80$ & $99$ & $228$ & $359$ & $769$ & $779$ & $4693$ \\
\midrule
$\gamma_c$ & $0.80$ & $0.90$ & $0.98$ & $0.98$ & $1.01$ & $1.00$ & $1.25$ \\
$\gamma_c$ fixed & $\mathbf{0.97}$ & $\mathbf{1.00}$ & $\mathbf{1.02}$ & $\mathbf{0.98}$ & $\mathbf{0.99}$ & $\mathbf{0.99}$ & $\mathbf{1.03}$ \\
\bottomrule
\end{tabular}
\end{table}

\noindent
Holding the argmin at its observed assignment puts the ratio within $0.03$ of unity over the
whole range with no free parameter. The two endpoints of the first row move with the assignment
rather than with the exponent.

\paragraph{Why the endpoints move.}
Equation~(\ref{eq:expand}) expands the norm for a fixed class, and the score takes a minimum over
all of them. A bootstrap replicate can therefore change which centroid wins. The share of
replicates that switch runs from $26\%$ at $n_c = 80$ down to $7\%$ at $n_c = 4693$, and the
switch compresses $T$ in the small classes while it inflates $T$ in the majority class. The two
rows of Table~\ref{tab:gamma} differ by that effect alone.

\section{The Reference Count}
\label{app:count}

The count enters Eq.~(\ref{eq:main-full}) as a divisor, so it can be read only where classes differ
in $n_c$. A balanced reference set makes $n_c \equiv \bar{n}$ and the substitution below into the
identity, which is why CIFAR-10, CIFAR-100 and ImageNet-1k are silent on this claim rather
than evidence against it. We therefore seek imbalance twice. DermaMNIST carries it at
clinical frequencies. ImageNet-1k receives it by construction in Appendix~\ref{app:nccontrol}, with
the count assigned at random across 1000 classes so that it cannot stand for the class.

Table~\ref{tab:count} isolates the contribution of $n_c$ under the clinical imbalance. Replacing
$n_c$ by the mean count $\bar{n} = N/C$ leaves every other quantity untouched, including
$\sigma_t(u)$, so the two models differ only in whether a class with $4693$ reference points is
treated differently from one with $80$. The gain from the count grows with the imbalance, and at
$58.7\times$ the count contributes $0.647$ of the total $R^2$ of $0.923$, or $70\%$.

\begin{table}[htbp]
\small
\centering
\caption{The reference count is not a refinement. Replacing $n_c$ in Eq.~(\ref{eq:main-full}) by
the mean count $\bar{n} = N/C$ leaves every other quantity untouched, including $\sigma_t(u)$.
The gain grows with the imbalance. The balanced column is ImageNet-1k, where $n_c \equiv \bar{n}$
makes the substitution the identity and the gain exactly zero. Every balanced setting we run reads
the same way, so the count is neither confirmed nor refuted there.}
\label{tab:count}
\begin{tabular}{lcccc}
\toprule
Imbalance           & $1\times$ & $6.1\times$ & $20.6\times$ & $58.7\times$ \\
                    & (balanced) & ($C{=}3$)   & ($C{=}5$)    & ($C{=}7$) \\
\midrule
$R^2$ with $n_c$    & $0.910$ & $\mathbf{0.974}$ & $\mathbf{0.973}$ & $\mathbf{0.923}$ \\
$R^2$ with $\bar{n}$ & $0.910$ & $0.756$ & $0.483$ & $0.276$ \\
\midrule
Gain from the count & $+0.000$ & $+0.218$ & $+0.490$ & $\mathbf{+0.647}$ \\
\bottomrule
\end{tabular}
\end{table}

\section{The 11 Scores in Full}
\label{app:eleven}

Figure~\ref{fig:eleven} places each of the 11 post-hoc scores by its two correlations but does
not print the values. Table~\ref{tab:eleven} supplies them for CIFAR-100 and for ImageNet-1k, and
adds the transverse dispersion $\sigma_t$, which the plot leaves out. The scores fall into three families by what they estimate,
and the sign of each correlation follows the family rather than the individual score. The two
dispersion scores carry the positive sign a practitioner expects while the distance and logit
families carry the negative one, so 9 of 11 are anti-correlated with the reliability of the
verdict they produce. The partition survives the move to $C = 1000$ with the same 9 scores on the
negative side. Every point agrees in the sign of both axes in both settings, which is the rule of
Eq.~(\ref{eq:rule}) read off the raw correlations. Appendix~\ref{app:odin} states how we compute
the logit family and what \texttt{odin} does.

\begin{table*}[htbp]
\small
\centering
\caption{11 post-hoc scores on CIFAR-100 and on ImageNet-1k. Spearman correlations against the
transverse dispersion, the plug-in estimate and the bootstrap instability, centered within query
group. Within each setting the second and third columns are the two sides of Eq.~(\ref{eq:rule}) and
Agree records whether they share a sign. The CIFAR-100 block gives the coordinates plotted in
Figure~\ref{fig:eleven}. 9 of 11 are anti-correlated with the reliability of the verdict they
produce in both settings, and Eq.~(\ref{eq:rule}) holds for all 11 in both.}
\label{tab:eleven}
\resizebox{\textwidth}{!}{%
\begin{tabular}{ll rrrc c rrrc}
\toprule
& & \multicolumn{4}{c}{\textit{CIFAR-100} ($C{=}50$)} && \multicolumn{4}{c}{\textit{ImageNet-1k} ($C{=}1000$)} \\
\cmidrule(lr){3-6}\cmidrule(lr){8-11}
Family & Score & $\rho(M, \sigma_t)$ & $\rho(M, \widehat{T})$ & $\rho(M, T)$ & Agree && $\rho(M, \sigma_t)$ & $\rho(M, \widehat{T})$ & $\rho(M, T)$ & Agree \\
\midrule
\multirow{2}{*}{Dispersion}
  & \texttt{knn\_std}    & $+0.767$ & $+0.734$ & $+0.695$ & \checkmark && $+0.652$ & $+0.652$ & $+0.654$ & \checkmark \\
  & \texttt{lid}         & $+0.532$ & $+0.509$ & $+0.483$ & \checkmark && $+0.512$ & $+0.512$ & $+0.510$ & \checkmark \\
\midrule
\multirow{4}{*}{Distance}
  & \texttt{d\_cls}      & $-0.202$ & $-0.309$ & $-0.334$ & \checkmark && $-0.167$ & $-0.167$ & $-0.239$ & \checkmark \\
  & \texttt{knn}         & $-0.419$ & $-0.495$ & $-0.481$ & \checkmark && $-0.405$ & $-0.405$ & $-0.439$ & \checkmark \\
  & \texttt{maha}        & $-0.265$ & $-0.390$ & $-0.355$ & \checkmark && $-0.150$ & $-0.151$ & $-0.193$ & \checkmark \\
  & \texttt{vim}         & $-0.324$ & $-0.317$ & $-0.371$ & \checkmark && $-0.187$ & $-0.187$ & $-0.203$ & \checkmark \\
\midrule
\multirow{5}{*}{Logit}
  & \texttt{energy}      & $-0.248$ & $-0.182$ & $-0.262$ & \checkmark && $-0.203$ & $-0.203$ & $-0.223$ & \checkmark \\
  & \texttt{maxlogit}    & $-0.248$ & $-0.183$ & $-0.263$ & \checkmark && $-0.200$ & $-0.200$ & $-0.220$ & \checkmark \\
  & \texttt{odin}        & $-0.153$ & $-0.123$ & $-0.179$ & \checkmark && $-0.079$ & $-0.079$ & $-0.087$ & \checkmark \\
  & \texttt{msp}         & $-0.167$ & $-0.130$ & $-0.184$ & \checkmark && $-0.059$ & $-0.059$ & $-0.074$ & \checkmark \\
  & \texttt{entropy}     & $-0.182$ & $-0.138$ & $-0.197$ & \checkmark && $-0.071$ & $-0.071$ & $-0.087$ & \checkmark \\
\bottomrule
\end{tabular}}
\end{table*}

\noindent
The dispersion column is the one \S\ref{sec:why} reasons about, and it is not the input the rule
takes. In both settings the two coincide closely enough to hide the difference. Every class holds
$n_c = 400$ reference points, so $\widehat{T}$ departs from $\sigma_t$ only through the penalty
channel, and on ImageNet-1k the hinge almost never fires so the two agree to three decimals. All 11
scores therefore agree in sign across the two columns in each block.

\section{How We Compute the Logit Family}
\label{app:odin}

The frozen backbone carries no classification head, so we fit a multinomial logistic probe on
$\mathcal{R}$ itself and read the logit family off that probe. All five of \texttt{energy},
\texttt{msp}, \texttt{maxlogit}, \texttt{entropy} and \texttt{odin} use it.

\texttt{odin} applies both components that \citet{liang2018enhancing} describe. Temperature scaling
runs at $1000$ and input perturbation at $\varepsilon = 0.0014$, which are the published
values. We tune neither. The original recipe selects $\varepsilon$ on a labeled out-of-distribution
validation set and that would contradict the label-free protocol of \S\ref{sec:setup}, so we treat
$\varepsilon$ exactly as we treat the neighbor count of \texttt{knn} or the subspace dimension of
\texttt{vim}.

Perturbation acts on the normalized input tensor, which makes \texttt{odin} the one score we
compute from raw pixels rather than from a cached embedding. Its embedding therefore depends on
$\mathcal{R}$ through the probe that supplies the gradient, while the other 10 scores read a fixed
embedding. Table~\ref{tab:odineps} sweeps $\varepsilon$ across the range
\citet{liang2018enhancing} consider. On CIFAR-100 the sign of $\rho(M, T)$ survives every value and
the top row of Table~\ref{tab:grid} settles at its reported value for every $\varepsilon \ge 0.001$.
On ImageNet-1k the correlation decays monotonically with the perturbation and reaches $+0.002$ at
the largest $\varepsilon$ the original recipe considers, so the sign holds at the published default
and is erased four times further out. Eq.~(\ref{eq:rule}) holds at every $\varepsilon$ in both
settings, because the predictor tracks the score down to zero alongside it.

\begin{table}[htbp]
\small
\centering
\caption{What the perturbation magnitude decides. Spearman correlations of \texttt{odin} centered
within query group, beside the $\alpha = 0$ row of Table~\ref{tab:grid}, which counts over all 6
settings and moves with $\varepsilon$ only through \texttt{odin} itself. The published default is
$\varepsilon = 0.0014$ (bold) and $\varepsilon = 0$ recovers temperature scaling alone.}
\label{tab:odineps}
\begin{tabular}{lrrcrrcr}
\toprule
& \multicolumn{2}{c}{\textit{CIFAR-100}} && \multicolumn{2}{c}{\textit{ImageNet-1k}} & \\
\cmidrule(lr){2-3}\cmidrule(lr){5-6}
$\varepsilon$ & $\rho(M, \sigma_t)$ & $\rho(M, T)$ && $\rho(M, \sigma_t)$ & $\rho(M, T)$
              && $\alpha = 0$ row \\
\midrule
$0$ & $-0.257$ & $-0.273$ && $-0.200$ & $-0.221$ && $53$ \\
$0.0005$ & $-0.202$ & $-0.227$ && $-0.144$ & $-0.159$ && $53$ \\
$0.001$ & $-0.167$ & $-0.194$ && $-0.103$ & $-0.115$ && $55$ \\
$\mathbf{0.0014}$ & $\mathbf{-0.153}$ & $\mathbf{-0.179}$ && $\mathbf{-0.079}$ & $\mathbf{-0.087}$ && $\mathbf{55}$ \\
$0.002$ & $-0.142$ & $-0.165$ && $-0.051$ & $-0.055$ && $55$ \\
$0.004$ & $-0.123$ & $-0.142$ && $-0.003$ & $+0.002$ && $54$ \\
\bottomrule
\end{tabular}
\end{table}

\section{Falsifying the Sign}
\label{app:falsification}

The negative sign of the distance family is a property of the embedding rather than of the theory,
and \S\ref{sec:rule} argues it can be switched off by destroying the anisotropy. Whitening does so
but changes the metric as well, so we isolate the mechanism by conditioning instead. Stratifying on
the alignment of the deviation with the leading principal directions, and leaving the metric and
$\sigma_t$ untouched, takes $\rho(r, \sigma_t)$ pooled over all query groups from $-0.306$ to $+0.019$ on CIFAR-100, from
$-0.223$ to $-0.045$ on ImageNet-1k and from $-0.831$ to $-0.007$ on DermaMNIST. The allocation is the mechanism
and the anisotropy is its cause.

\begin{table}[htbp]
\small
\centering
\caption{The sign against the alignment it is conditioned on. Entries are $\rho(r, \sigma_t)$ after
stratifying on the deviation's share of the leading $P$ principal directions, in $20$ bins.
Uncontrolled the correlation is $-0.306$ on CIFAR-100, $-0.223$ on ImageNet-1k and $-0.831$ on
DermaMNIST, which are the pooled entries of Table~\ref{tab:groups}. Conditioning removes it in all
three.}
\label{tab:falsify}
\begin{tabular}{lccc} \toprule $P$ & \textit{CIFAR-100} & \textit{ImageNet-1k} & \textit{DermaMNIST}
\\
\midrule
$1$ & $-0.255$ & $-0.045$ & $-0.572$ \\
$5$ & $-0.116$ & $+0.037$ & $-0.463$ \\
$10$ & $+0.019$ & $+0.132$ & $-0.072$ \\
$20$ & $+0.040$ & $+0.190$ & $-0.218$ \\
$50$ & $+0.054$ & $+0.207$ & $-0.075$ \\
$100$ & $+0.044$ & $+0.210$ & $-0.007$ \\
\bottomrule
\end{tabular}
\end{table}

\noindent
The participation ratio of the class scatter is $45.5$ on CIFAR-100, $34.3$ on ImageNet-1k and
$18.3$ on DermaMNIST, so a given $P$ covers more of the spectrum as the clouds flatten. The two
natural-image settings cross zero and then overshoot, since conditioning past the crossing removes
more than the mechanism, and we read the intervention at the crossing rather than at the largest
$P$. DermaMNIST starts four times further from zero and approaches it from below, so its reading is
the largest $P$.

\section{A Representation Check}
\label{app:backbone}

The intervention above reshapes the embedding we have. A coarser one on the same axis is to build
the embedding differently, and we use it to check that Eq.~(\ref{eq:rule}) still calls the sign.
We re-encode every dataset three times. The first encoder is CLIP \citep{radford2021learning},
trained on a different objective with the architecture, the embedding dimension, the input resolution
and the patch size held at the values our backbone uses. The second is ResNet-50 \citep{he2016deep},
a convolutional network trained with labels, which moves the architecture family, the embedding
dimension and the supervision signal with it. The third is DINOv3 \citep{simeoni2025dinov3}, a
recent self-supervised objective trained on a far larger corpus. It shares our patch size, our
embedding dimension and our normalization constants, so its input tensor is identical to ours to the
last bit. We pin the crop geometry of the second and third to ours, so the same crop enters all four
encoders. Every hyperparameter of
the detector is transplanted unchanged and nothing is reselected, so neither check introduces a free
parameter. The reference sets, the query sets and the bootstrap draws are the same images in the
same order.

Each check therefore moves as little as it can. None moves the input geometry, which stays at
$224 \times 224$ so that the same crop of the same image reaches every encoder. An encoder at a different resolution or patch size would confound the representation with how much of the image the detector sees. 

Run on our own embedding the same code path returns all 66 entries of Table~\ref{tab:eleven} to 3 decimals and recovers the grid of
Table~\ref{tab:grid} in each of the 6 settings, so the numbers in
Tables~\ref{tab:repcheckfull}, \ref{tab:repcheckconv} and~\ref{tab:repcheckdinov3} differ from our
own only by the encoding. Equation~(\ref{eq:main-full}) is undamaged by any of the three changes.
Across the 18 fits the median $T/\widehat{T}$ lies between $0.91$ and $0.99$ and the $R^2$ between
$0.66$ and $0.97$.

The sign of $\rho(M,T)$ is unchanged from the one our own embedding gives in 63 of the 66
score-and-setting pairs under CLIP, in 62 under ResNet-50 and in 64 under DINOv3, and the rule calls
it in 63, in 65 and in all 66. Eq.~(\ref{eq:rule}) never fails under DINOv3, whose input tensor matches ours exactly, so the rule
follows the geometry rather than a particular checkpoint. The logit family carries the positive sign under
imbalance on all four encodings, so the reversal of \S\ref{sec:derma-flip} travels with the class prior rather than with the
representation. The three tables share a format. Each reports $\rho(M,T)$ centered within query
group. A dagger marks a cell where Eq.~(\ref{eq:rule}) reads the sign backwards. The first summary
row counts the scores whose sign matches the one our own embedding gives and the second those
satisfying Eq.~(\ref{eq:rule}).

\begin{table}[htbp]
\small
\centering
\caption{The CLIP re-encoding.}
\label{tab:repcheckfull}
\begin{tabular}{llrrrrrr}
\toprule
Family & Score & \textit{CIFAR-100} & \textit{CIFAR-10} & \textit{ImageNet-1k} & \textit{Derma} $C{=}3$ & \textit{Derma} $C{=}5$ & \textit{Derma} $C{=}7$ \\
\midrule
\multirow{2}{*}{Dispersion}
  & \texttt{knn\_std} & $+0.787$ & $+0.790$ & $+0.716$ & $+0.528$ & $+0.253$ & $+0.372$ \\
  & \texttt{lid} & $+0.570$ & $+0.536$ & $+0.521$ & $+0.479$ & $+0.510$ & $+0.550$ \\
\midrule
\multirow{4}{*}{Distance}
  & \texttt{d\_cls} & $-0.331$ & $-0.536$ & $-0.180$ & $-0.500$ & $-0.458$ & $-0.213$ \\
  & \texttt{knn} & $-0.511$ & $-0.681$ & $-0.313$ & $-0.568$ & $-0.532$ & $-0.387$ \\
  & \texttt{maha} & $-0.443$ & $-0.660$ & $-0.219$ & $-0.560$ & $-0.505$ & $-0.388$ \\
  & \texttt{vim} & $-0.422$ & $-0.364$ & $-0.151$ & $-0.327$ & $-0.323$ & $-0.294$ \\
\midrule
\multirow{5}{*}{Logit}
  & \texttt{energy} & $-0.074^{\dagger}$ & $+0.056$ & $-0.090$ & $+0.232$ & $+0.294$ & $+0.218$ \\
  & \texttt{maxlogit} & $-0.085^{\dagger}$ & $+0.046$ & $-0.097$ & $+0.232$ & $+0.290$ & $+0.204$ \\
  & \texttt{odin} & $-0.095$ & $-0.185$ & $-0.063$ & $+0.270$ & $+0.369$ & $+0.229$ \\
  & \texttt{msp} & $-0.122$ & $-0.032$ & $-0.119$ & $+0.154$ & $+0.153$ & $+0.009$ \\
  & \texttt{entropy} & $-0.123$ & $-0.026^{\dagger}$ & $-0.130$ & $+0.171$ & $+0.173$ & $+0.017$ \\
\midrule
\multicolumn{2}{l}{Sign unchanged from ours} & $\mathbf{11}$ / $11$ & $9$ / $11$ & $\mathbf{11}$ / $11$ & $\mathbf{11}$ / $11$ & $\mathbf{11}$ / $11$ & $10$ / $11$ \\
\multicolumn{2}{l}{Eq.~(\ref{eq:rule}) holds} & $9$ / $11$ & $10$ / $11$ & $\mathbf{11}$ / $11$ & $\mathbf{11}$ / $11$ & $\mathbf{11}$ / $11$ & $\mathbf{11}$ / $11$ \\
\bottomrule
\end{tabular}
\end{table}

\begin{table}[htbp]
\small
\centering
\caption{The ResNet-50 re-encoding.}
\label{tab:repcheckconv}
\begin{tabular}{llrrrrrr}
\toprule
Family & Score & \textit{CIFAR-100} & \textit{CIFAR-10} & \textit{ImageNet-1k} & \textit{Derma} $C{=}3$ & \textit{Derma} $C{=}5$ & \textit{Derma} $C{=}7$ \\
\midrule
\multirow{2}{*}{Dispersion}
  & \texttt{knn\_std} & $+0.656$ & $+0.566$ & $+0.617$ & $+0.391$ & $+0.248$ & $+0.180$ \\
  & \texttt{lid} & $+0.474$ & $+0.350$ & $+0.304$ & $+0.427$ & $+0.471$ & $+0.352$ \\
\midrule
\multirow{4}{*}{Distance}
  & \texttt{d\_cls} & $-0.181$ & $-0.454$ & $+0.027$ & $-0.457$ & $-0.533$ & $-0.009$ \\
  & \texttt{knn} & $-0.334$ & $-0.570$ & $+0.003$ & $-0.431$ & $-0.524$ & $-0.014$ \\
  & \texttt{maha} & $-0.343$ & $-0.589$ & $+0.001$ & $-0.456$ & $-0.521$ & $+0.015^{\dagger}$ \\
  & \texttt{vim} & $-0.361$ & $-0.368$ & $-0.144$ & $-0.173$ & $-0.274$ & $+0.141$ \\
\midrule
\multirow{5}{*}{Logit}
  & \texttt{energy} & $-0.209$ & $-0.100$ & $-0.178$ & $+0.217$ & $+0.311$ & $+0.224$ \\
  & \texttt{maxlogit} & $-0.218$ & $-0.092$ & $-0.186$ & $+0.221$ & $+0.305$ & $+0.209$ \\
  & \texttt{odin} & $-0.126$ & $-0.074$ & $-0.264$ & $+0.172$ & $+0.256$ & $+0.265$ \\
  & \texttt{msp} & $-0.211$ & $-0.066$ & $-0.168$ & $+0.195$ & $+0.213$ & $+0.100$ \\
  & \texttt{entropy} & $-0.213$ & $-0.082$ & $-0.180$ & $+0.210$ & $+0.254$ & $+0.131$ \\
\midrule
\multicolumn{2}{l}{Sign unchanged from ours} & $\mathbf{11}$ / $11$ & $\mathbf{11}$ / $11$ & $8$ / $11$ & $\mathbf{11}$ / $11$ & $\mathbf{11}$ / $11$ & $10$ / $11$ \\
\multicolumn{2}{l}{Eq.~(\ref{eq:rule}) holds} & $\mathbf{11}$ / $11$ & $\mathbf{11}$ / $11$ & $\mathbf{11}$ / $11$ & $\mathbf{11}$ / $11$ & $\mathbf{11}$ / $11$ & $10$ / $11$ \\
\bottomrule
\end{tabular}
\end{table}

\begin{table}[htbp]
\small
\centering
\caption{The DINOv3 re-encoding.}
\label{tab:repcheckdinov3}
\begin{tabular}{llrrrrrr}
\toprule
Family & Score & \textit{CIFAR-100} & \textit{CIFAR-10} & \textit{ImageNet-1k} & \textit{Derma} $C{=}3$ & \textit{Derma} $C{=}5$ & \textit{Derma} $C{=}7$ \\
\midrule
\multirow{2}{*}{Dispersion}
  & \texttt{knn\_std} & $+0.743$ & $+0.737$ & $+0.821$ & $+0.485$ & $+0.427$ & $+0.427$ \\
  & \texttt{lid} & $+0.541$ & $+0.503$ & $+0.565$ & $+0.364$ & $+0.366$ & $+0.450$ \\
\midrule
\multirow{4}{*}{Distance}
  & \texttt{d\_cls} & $-0.226$ & $-0.346$ & $+0.008$ & $-0.374$ & $-0.398$ & $-0.279$ \\
  & \texttt{knn} & $-0.351$ & $-0.385$ & $-0.244$ & $-0.394$ & $-0.409$ & $-0.408$ \\
  & \texttt{maha} & $-0.345$ & $-0.549$ & $-0.015$ & $-0.448$ & $-0.460$ & $-0.402$ \\
  & \texttt{vim} & $-0.409$ & $-0.519$ & $-0.209$ & $-0.166$ & $-0.260$ & $-0.282$ \\
\midrule
\multirow{5}{*}{Logit}
  & \texttt{energy} & $-0.301$ & $-0.265$ & $-0.397$ & $+0.132$ & $+0.142$ & $+0.109$ \\
  & \texttt{maxlogit} & $-0.307$ & $-0.278$ & $-0.396$ & $+0.135$ & $+0.141$ & $+0.104$ \\
  & \texttt{odin} & $-0.251$ & $-0.294$ & $-0.323$ & $+0.136$ & $+0.138$ & $+0.065$ \\
  & \texttt{msp} & $-0.271$ & $-0.291$ & $-0.286$ & $+0.117$ & $+0.100$ & $+0.030$ \\
  & \texttt{entropy} & $-0.288$ & $-0.303$ & $-0.319$ & $+0.129$ & $+0.112$ & $+0.043$ \\
\midrule
\multicolumn{2}{l}{Sign unchanged from ours} & $\mathbf{11}$ / $11$ & $\mathbf{11}$ / $11$ & $10$ / $11$ & $\mathbf{11}$ / $11$ & $\mathbf{11}$ / $11$ & $10$ / $11$ \\
\multicolumn{2}{l}{Eq.~(\ref{eq:rule}) holds} & $\mathbf{11}$ / $11$ & $\mathbf{11}$ / $11$ & $\mathbf{11}$ / $11$ & $\mathbf{11}$ / $11$ & $\mathbf{11}$ / $11$ & $\mathbf{11}$ / $11$ \\
\bottomrule
\end{tabular}
\end{table}

\noindent
The daggered cells all sit at $\lvert \rho(M,\widehat{T}) \rvert < 0.03$, so in none of them does
the predictor carry a sign for the rule to read backwards. Where the sign does differ from ours the
rule follows the re-encoding rather than our own embedding. In the 3 settings that identify the reference count it calls all 33 under CLIP, 32 of 33 under
ResNet-50 and all 33 under DINOv3.

\section{The Geometry Across Settings}
\label{app:imagenet}

Table~\ref{tab:groups} reads the mechanism off all three settings at once. The allocation effective
dimension rises from the in-distribution group to the far sources in each, and $\sigma_t$ falls with
it. ImageNet-1k has no held-out classes, so the OpenOOD near pair stands where the other settings
put a held-out split.

\noindent
Two readings separate the settings and both belong to the embedding rather than to the theory. The
in-distribution group reverses on DermaMNIST alone. It reads $-0.200$ on CIFAR-100 and $-0.210$ on
ImageNet-1k against the $+0.15$ DermaMNIST gives, so the radius and the dispersion move together on DermaMNIST alone. The pooled correlation also weakens with the class count, from
$-0.83$ to $-0.306$ to $-0.223$. Every OpenOOD source is a natural image and none sits as far from
the reference cloud as SVHN or DTD sits from a $28$-pixel lesion. The sign of the distance family
survives both readings, and Eq.~(\ref{eq:rule}) calls it for all 11 scores in every setting.

\section{Imbalance by Construction}
\label{app:nccontrol}

A clinical frequency ties the count to the class. A rare lesion has few reference points and also
looks different from a common one, and no observational reference set can separate the two. ImageNet
can, because a pool of 500 images per class lets us decide which class receives which count.
Assigning the counts at random makes the reference count independent of class identity by
construction. All four experiments below use the 1000 classes of ImageNet-1k and the plug-in estimate
of Eq.~(\ref{eq:main-full}). None of them needs the logit family, so none fits a probe.

Table~\ref{tab:ncsweep} draws $n_c$ log-uniformly at a target imbalance and repeats the
substitution of \S\ref{sec:scaling}. The balanced control returns exactly zero, as it must when
$n_c \equiv \bar{n}$. The gain then grows with the imbalance as it does on DermaMNIST.

\begin{table}[htbp]
\small
\centering
\caption{The reference count under random assignment. $n_c$ is drawn log-uniformly at the target
imbalance and assigned to classes independently of their identity. Every other quantity is left as
it was. The DermaMNIST column is Table~\ref{tab:count}.}
\label{tab:ncsweep}
\begin{tabular}{lccccc}
\toprule
Target imbalance & $1\times$ & $6.1\times$ & $20.6\times$ & $58.7\times$ & \textit{Derma} \\
\midrule
$n_c$ range & $500$--$500$ & $82$--$499$ & $24$--$500$ & $9$--$496$ & $80$--$4693$ \\
\midrule
$R^2$ with $n_c$ & $0.905$ & $0.891$ & $0.858$ & $0.776$ & $0.923$ \\
$R^2$ with $\bar{n}$ & $0.905$ & $0.545$ & $0.319$ & $0.261$ & $0.276$ \\
\midrule
Gain from the count & $+0.000$ & $+0.346$ & $+0.539$ & $\mathbf{+0.515}$ & $+0.647$ \\
\bottomrule
\end{tabular}
\end{table}

\noindent
The decisive test holds the multiset of counts fixed and permutes only which class receives which.
If class identity were doing the work, breaking the correspondence would collapse the gain toward
zero. Over 5 permutations of the $58.7\times$ counts the gain is $+0.524 \pm 0.068$ and never falls
below $+0.433$. The spread it does carry comes from which classes happen to receive the smallest
counts, where $\sigma_t$ is noisiest. No permutation moves the gain toward zero.

ImageNet also allows a test that DermaMNIST forbids outright. Fit the same class twice at two different counts, with its identity held completely fixed.
Table~\ref{tab:ncwithin} does that. The predicted ratio is $\sqrt{n_{\text{hi}}/n_{\text{lo}}}$ and
carries no free parameter.

\begin{table}[htbp]
\small
\centering
\caption{The exponent with class identity held fixed. Each row fits every class twice at $n_c$ high
and at $n_c$ low, then reports the median over classes of
$\overline{T}_c(\text{lo})/\overline{T}_c(\text{hi})$.}
\label{tab:ncwithin}
\begin{tabular}{lcccc}
\toprule
$n_c$ high $\to$ low & Predicted & Observed median & $10$--$90\%$ & Classes \\
\midrule
$400 \to 50$ & $2.828$ & $\mathbf{2.725}$ & $2.453$--$3.003$ & $994$ \\
$400 \to 100$ & $2.000$ & $\mathbf{1.959}$ & $1.802$--$2.123$ & $994$ \\
$500 \to 25$ & $4.472$ & $\mathbf{4.201}$ & $3.627$--$4.833$ & $994$ \\
\bottomrule
\end{tabular}
\end{table}

\noindent
All three fall a little short of the prediction and fall further as the ratio widens.
Proposition~\ref{prop:min} accounts for the gap. A smaller class carries a larger fluctuation and a
narrower gap to its runner-up, so the minimum compresses its spread, and the compression grows with
the ratio. Reading $\gamma_c = \overline{T}_c\sqrt{n_c}/\overline{\sigma}_{t,c}$ off the
random assignment and holding the argmin at its observed value puts the ratio within $0.008$ of unity
for every bin above $n_c = 30$.

\begin{table}[htbp]
\small
\centering
\caption{$\gamma_c$ under random assignment, binned by the count the class was given. The second row
holds the argmin at its observed assignment, as the second row of Table~\ref{tab:gamma} does.}
\label{tab:ncgamma}
\begin{tabular}{lcccccc}
\toprule
$n_c$ bin & $5-14$ & $15-29$ & $30-59$ & $60-119$ & $120-249$ & $250-500$ \\
Classes & $116$ & $187$ & $192$ & $172$ & $174$ & $159$ \\
\midrule
$\gamma_c$ & $0.894$ & $0.903$ & $0.935$ & $0.960$ & $1.005$ & $1.031$ \\
$\gamma_c$ fixed & $1.081$ & $1.033$ & $\mathbf{1.006}$ & $\mathbf{0.997}$ & $\mathbf{0.997}$ & $\mathbf{0.998}$ \\
\bottomrule
\end{tabular}
\end{table}

\noindent
The two sources of imbalance answer different halves of one question. Random assignment shows that
the count acts causally. DermaMNIST shows that it acts in a reference set nobody designed. Both reach down to $n_c = 30$. Below that the ratio departs from unity and the fit at the widest
imbalance drops to $R^2 = 0.776$, which is the range \S\ref{sec:scaling} does not claim.

\section{The Ablation Behind the Rule}
\label{app:ablation}

Section~\ref{sec:rule} states that the rule of Eq.~(\ref{eq:rule}) uses the full model and that the count is essential. Here we weaken the predictor along its two non-trivial axes and count sign
agreements against the true instability. The axes are the exponent $\alpha$ on the reference count
and the weight $\lambda$ of the penalty channel. Writing the predictor as
$\sigma_t^2 / n_c^{\alpha} + \lambda^2 \operatorname{Var}[\mathrm{pen}]$, the model sits at
$\alpha = 1$ with $\lambda$ equal to the detector's own hinge weight. The square root of
Eq.~(\ref{eq:main-full}) is a monotone transform and leaves every rank correlation unchanged, so
these two axes are the only ones that matter.

\begin{table}[htbp]
\small
\centering
\caption{The count is what this grid identifies. Entries are sign agreements out of 66, over 6
settings and 11 scores, for the predictor
$\sigma_t^2/n_c^{\alpha} + \lambda^2 \operatorname{Var}[\mathrm{pen}]$. The model sits at
$\alpha = 1$ and $\lambda = 5$ (bold). Dropping the count entirely, the top row, costs 10 agreements
that no penalty weight recovers. At the derived exponent the penalty channel recovers 2 agreements.
The maximum is a plateau, so the grid identifies the count rather than the exponent or the penalty
weight.}
\label{tab:grid}
\begin{tabular}{lccccc}
\toprule
$\alpha \;\backslash\; \lambda$ & $0$ & $1$ & $2.5$ & $5$ & $10$ \\
\midrule
$0.00$ & $55$ & $55$ & $55$ & $55$ & $55$ \\
$0.25$ & $64$ & $64$ & $64$ & $65$ & $65$ \\
$0.50$ & $64$ & $64$ & $64$ & $64$ & $64$ \\
$0.75$ & $65$ & $65$ & $65$ & $65$ & $65$ \\
$1.00$ & $63$ & $63$ & $65$ & $\mathbf{65}$ & $63$ \\
$1.25$ & $63$ & $64$ & $65$ & $63$ & $58$ \\
\bottomrule
\end{tabular}
\end{table}

\noindent
Dropping the count entirely at $\alpha = 0$ costs 10 agreements out of 66, and no choice
of penalty weight recovers them. The three balanced settings contribute $33$ to every cell, because a balanced reference set makes
$n_c^{\alpha}$ a constant and the $\alpha$ axis cannot move a rank there. The whole of the
discrimination therefore comes from the three that carry imbalance. The model's own values reach the maximum of the grid and the
maximum is a plateau, so the rule is robust to the exact exponent and to the penalty weight alike.
The grid reads the sign of a rank correlation and is insensitive to the precise exponent by
construction, and the stronger claim that the exponent is one half is made where the model itself is
fit, in \S\ref{sec:scaling}. The count is essential for the same reason it is there. A logit
reflects the class prior, so under heavy imbalance it correlates with $n_c$, which instability
carries in its denominator. A score coupled to that denominator reaches $T$ without passing through
$\sigma_t$, so a predictor that omits the count reads the sign backwards. This is why the dispersion-only predictor at $\alpha = 0$ scores $55$ where the full model scores
$65$.

\section{The Stretch Channel}
\label{app:stretch}

Section~\ref{sec:related-geometry} puts the stretch channel $W$ of Eq.~(\ref{eq:stretch}) through
Eq.~(\ref{eq:rule}). The rule reads $\widehat{T}$, so Table~\ref{tab:stretch} reports
$\rho(W, \widehat{T})$ beside $\rho(W, T)$. Here $\delta = z - \hat\mu_c$. The size channel and the product of the two, which is the Mahalanobis score itself, run alongside
it. The Cauchy--Schwarz bound of \S\ref{sec:related-geometry} holds in every query of every setting,
and $\rho(\log W, -\log u^{\top}\Sigma u)$ reads $0.862$ on CIFAR-100, $0.828$ on ImageNet-1k and $0.931$
on DermaMNIST.

\begin{table}[htbp]
\small
\centering
\caption{The three channels of the Mahalanobis decomposition against instability. Spearman
correlations, centered within query group, as means over the 10 seeds of
Appendix~\ref{app:seeds}. The two columns of each dataset are the two sides of
Eq.~(\ref{eq:rule}). Every entry is negative except the stretch channel on ImageNet-1k, where both sides sit within
$0.03$ of zero and the predictor carries no sign to read. Elsewhere all three channels rank the most
reproducible verdicts as the most suspicious.}
\label{tab:stretch}
\begin{tabular}{l rr c rr c rr}
\toprule
& \multicolumn{2}{c}{\textit{CIFAR-100}} && \multicolumn{2}{c}{\textit{ImageNet-1k}}
&& \multicolumn{2}{c}{\textit{DermaMNIST}} \\
\cmidrule(lr){2-3}\cmidrule(lr){5-6}\cmidrule(lr){8-9}
Channel & $\rho(M,\widehat{T})$ & $\rho(M,T)$ && $\rho(M,\widehat{T})$ & $\rho(M,T)$
        && $\rho(M,\widehat{T})$ & $\rho(M,T)$ \\
\midrule
Stretch $W$                              & $-0.293$ & $-0.184$
  && $-0.023$ & $+0.014$
  && $\mathbf{-0.271}$ & $\mathbf{-0.231}$ \\
Size $\lVert \delta \rVert^2$             & $-0.299$ & $-0.325$
  && $-0.151$ & $-0.228$
  && $-0.403$ & $-0.395$ \\
Product $\delta^{\top}\Sigma^{-1}\delta$ & $-0.388$ & $-0.347$
  && $-0.139$ & $-0.197$
  && $-0.394$ & $-0.369$ \\
\bottomrule
\end{tabular}
\end{table}

\noindent
The prediction of \S\ref{sec:related-geometry} is a statement about deferral, so we measure it
with the instrument of \S\ref{sec:downstream} as well. Table~\ref{tab:stretchaurc} defers on each
channel and compares the area against the random baseline. It follows the pooled convention of
Appendix~\ref{app:protocol}, which is the one Table~\ref{tab:worse} uses.

\begin{table*}[htbp]
\small
\centering
\caption{Deferring by any channel of the Mahalanobis decomposition is worse than deferring at
random. AURC and its position $\Delta$ against the random baseline, pooled and uncentered, as means
over the 10 seeds of Appendix~\ref{app:seeds} with a $95\%$ interval on $\Delta$. Lower AURC is
better and a positive $\Delta$ is a loss. The sign is unanimous in all 10 seeds in all 9 cells.}
\label{tab:stretchaurc}
\resizebox{\textwidth}{!}{%
\begin{tabular}{l rrl c rrl c rrl}
\toprule
& \multicolumn{3}{c}{\textit{CIFAR-100}} && \multicolumn{3}{c}{\textit{ImageNet-1k}}
&& \multicolumn{3}{c}{\textit{DermaMNIST}} \\
\cmidrule(lr){2-4}\cmidrule(lr){6-8}\cmidrule(lr){10-12}
Channel & $\rho(M,\widehat{T})$ & AURC & $\Delta$ (\%) && $\rho(M,\widehat{T})$ & AURC & $\Delta$ (\%)
        && $\rho(M,\widehat{T})$ & AURC & $\Delta$ (\%) \\
\midrule
Stretch $W$                              & $-0.405$ & 0.1323 & $\mathbf{+2.44}${\scriptsize$\,\pm0.09$}
  && $-0.087$ & 0.1523 & $\mathbf{+0.87}${\scriptsize$\,\pm0.18$}
  && $-0.663$ & 0.0701 & $\mathbf{+12.75}${\scriptsize$\,\pm0.14$} \\
Size $\lVert \delta \rVert^2$             & $-0.451$ & 0.1342 & $+3.94${\scriptsize$\,\pm0.11$}
  && $-0.207$ & 0.1565 & $\mathbf{+3.71}${\scriptsize$\,\pm0.15$}
  && $-0.693$ & 0.0703 & $+13.06${\scriptsize$\,\pm0.18$} \\
Product $\delta^{\top}\Sigma^{-1}\delta$ & $-0.537$ & 0.1343 & $+4.03${\scriptsize$\,\pm0.10$}
  && $-0.198$ & 0.1557 & $\mathbf{+3.16}${\scriptsize$\,\pm0.15$}
  && $-0.696$ & 0.0704 & $+13.09${\scriptsize$\,\pm0.16$} \\
\midrule
\textit{random} & \textit{$+0.000$} & \textit{0.1291} & \textit{---}
  && \textit{$+0.000$} & \textit{0.1509} & \textit{---}
  && \textit{$+0.000$} & \textit{0.0622} & \textit{---} \\
\bottomrule
\end{tabular}}
\end{table*}

\noindent
The product row is the Mahalanobis score written in the notation of the decomposition, and it
tracks the \texttt{maha} row of Table~\ref{tab:worse} on both datasets. The stretch channel is the
one \citet{janiak2026geometry} shape representations to control, and of the three it is the one
that costs the least, at $2.4\%$ on CIFAR-100 against the $3.9\%$ of the size channel it
compensates. Controlling it buys detection performance and sells the stability of the verdict.

\section{The Instability--Coverage Curves}
\label{app:iccurve}

\begin{figure*}[htbp]
\centering
\includegraphics[width=\textwidth]{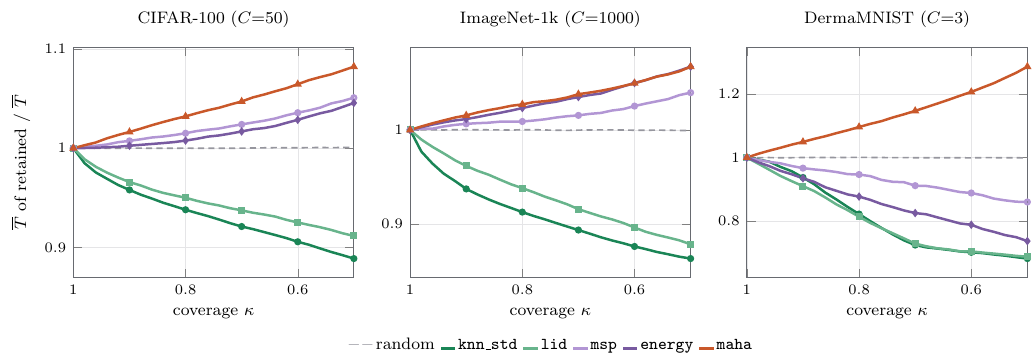}
\caption{Instability against coverage. The random baseline is flat at one by construction, and a
curve above it retains a population whose verdicts are less reproducible than the one it started
with. Mahalanobis rises in all three settings while the logit scores change side with the dataset.}
\label{fig:iccurve}
\end{figure*}

Figure~\ref{fig:iccurve} draws the curves that Table~\ref{tab:worse} summarizes. Each is the mean
verdict instability of the retained set as the coverage falls, divided by the mean over all queries
so that the random baseline sits at one in every panel. A score that buys stability bends its curve
down and a score that costs it bends the curve up. The curves come from a single seed, so the areas
differ from Table~\ref{tab:worse} by less than half a point.

\noindent
The rise is monotone wherever it happens, so the cost of a wrong-signed score grows with how much
of the population it defers. CIFAR-100 and ImageNet-1k give the same shape, and the imbalance of DermaMNIST moves the logit
scores below the baseline while leaving Mahalanobis above it.

\section{Seeds}
\label{app:seeds}

Table~\ref{tab:seeds} gives the seed-to-seed behavior of $\Delta$ behind Table~\ref{tab:worse}, and
Table~\ref{tab:level} separates the two quantities that a re-run moves. A seed redraws the reference
subsample, the query subsample, the $B$ bootstrap replicates and the random baseline
(Appendix~\ref{app:protocol}).

\begin{table*}[htbp]
\small
\centering
\caption{The relative position $\Delta$ of each score across 10 seeds, in percent. The sign is the
claim and it is unanimous in every cell. The spread within a score is an order of magnitude smaller
than its distance from the baseline.}
\label{tab:seeds}
\resizebox{\textwidth}{!}{%
\begin{tabular}{l rrrr c rrrr c rrrr}
\toprule
& \multicolumn{4}{c}{\textit{CIFAR-100}} && \multicolumn{4}{c}{\textit{ImageNet-1k}}
&& \multicolumn{4}{c}{\textit{DermaMNIST}} \\
\cmidrule(lr){2-5}\cmidrule(lr){7-10}\cmidrule(lr){12-15}
Score & mean & sd & range & sign && mean & sd & range & sign && mean & sd & range & sign \\
\midrule
\texttt{knn\_std} & $-6.99$ & $0.20$ & {\scriptsize$[-7.24,-6.67]$} & $10/10$ && $-8.82$ & $0.28$ & {\scriptsize$[-9.36,-8.41]$} & $10/10$ && $-19.70$ & $0.28$ & {\scriptsize$[-20.06,-19.30]$} & $10/10$ \\
\texttt{lid} & $-5.57$ & $0.22$ & {\scriptsize$[-5.84,-5.19]$} & $10/10$ && $-6.81$ & $0.21$ & {\scriptsize$[-7.13,-6.45]$} & $10/10$ && $-19.68$ & $0.27$ & {\scriptsize$[-20.09,-19.22]$} & $10/10$ \\
\texttt{energy} & $+1.95$ & $0.28$ & {\scriptsize$[+1.56,+2.50]$} & $10/10$ && $+3.00$ & $0.27$ & {\scriptsize$[+2.54,+3.39]$} & $10/10$ && $-14.26$ & $0.42$ & {\scriptsize$[-14.85,-13.46]$} & $10/10$ \\
\texttt{msp} & $+2.30$ & $0.16$ & {\scriptsize$[+2.13,+2.66]$} & $10/10$ && $+1.87$ & $0.22$ & {\scriptsize$[+1.59,+2.23]$} & $10/10$ && $-6.84$ & $0.53$ & {\scriptsize$[-7.59,-6.03]$} & $10/10$ \\
\texttt{maha} & $+3.93$ & $0.15$ & {\scriptsize$[+3.68,+4.20]$} & $10/10$ && $+3.19$ & $0.23$ & {\scriptsize$[+2.81,+3.61]$} & $10/10$ && $+13.09$ & $0.25$ & {\scriptsize$[+12.70,+13.49]$} & $10/10$ \\
\bottomrule
\end{tabular}}
\end{table*}

\noindent
The level and the contrast behave differently, which is why we report $\Delta$ rather than the area
itself. The baseline area matches its predicted value $\tfrac{1}{2}\overline{T}$ to 4 decimals in all three. That confirms it as a functional of the query set rather than an estimate drawn from
it. The level then moves by a few percent between seeds and carries every score with it at a correlation of $0.85$ or above in all 15 cells. The contrast holds still. The widest interval on $\Delta$ is $0.76$ percentage points against margins between $2$ and $20$.

\begin{table}[htbp]
\small
\centering
\caption{What a re-run moves. The level of the table drifts by a few percent and takes every row
with it. The position of a score relative to the baseline does not.}
\label{tab:level}
\begin{tabular}{lccc}
\toprule
 & \textit{CIFAR-100} & \textit{ImageNet-1k} & \textit{DermaMNIST} \\
\midrule
Baseline area $\mathrm{AURC}_{\mathrm{rand}}$ & 0.1291 & 0.1509 & 0.0622 \\
Predicted $\tfrac{1}{2}\overline{T}$ & 0.1291 & 0.1509 & 0.0622 \\
Level spread across seeds & $1.22\%$ & $2.19\%$ & $3.83\%$ \\
Widest interval on $\Delta$ & $0.41$\,pp & $0.34$\,pp & $0.76$\,pp \\
$\mathrm{corr}(\mathrm{AURC}_M, \mathrm{AURC}_{\mathrm{rand}})$ & $0.85$--$0.96$ & $0.89$--$0.95$ & $0.91$--$0.99$ \\
\bottomrule
\end{tabular}
\end{table}

\noindent
We evaluate the baseline as the mean of 20 uniform draws rather than one. A single draw is
itself an estimate, with a coefficient of variation of $0.13\%$ on CIFAR-100 and $0.55\%$ on the
smaller DermaMNIST query set, and that noise would propagate into every $\Delta$ in the column.
Averaging removes it. The result agrees with $\tfrac{1}{2}\overline{T}$, which carries no noise at
all.

The predictor of the rule is stable across the same seeds. The sign of $\rho(M,\widehat{T})$ is
unanimous over 10 seeds for all 5 scores on both datasets, including the smallest magnitude in
either table, energy on CIFAR-100 at $-0.108 \pm 0.014$. Eq.~(\ref{eq:rule}) holds in all 10 seeds
in all 10 cells.

\section{The Reference-Count Sweep}
\label{app:sweep}

Table~\ref{tab:sweep} sweeps the reference count on CIFAR-100 and reports per-score AURC at each
size. Every score stays on the same side of the random baseline throughout, while the mean verdict
instability at full coverage falls with an exponent of $0.45$. Holding the argmin fixed raises that
exponent to $0.50$, so the same competition that moves the endpoints of Table~\ref{tab:gamma} also
flattens the sweep. This is the evidence behind the small-sample claim of \S\ref{sec:worse}.

The margins move with the sweep while the sign holds. Read as $\Delta$, those of the logit
scores narrow as $n_c$ grows, from $+6.2\%$ to $+2.1\%$ for energy, while \texttt{maha} holds near
$+4\%$ and \texttt{knn\_std} near $-7\%$. The claim is the sign across the sizes we test.

\begin{table*}[htbp]
\small
\centering
\caption{Per-score AURC on CIFAR-100 across reference-set sizes $n_c$. Lower is better.
The two dispersion scores beat the random baseline at every size and the three distance and logit
scores below it fall short at every size, 50 cells over the two settings with no crossing, while the
mean instability shrinks with $n_c$. The $n_c = 400$ column is an independent run from
Table~\ref{tab:worse} and predates the seeded protocol of Appendix~\ref{app:seeds}. Its baseline
area sits $0.43\%$ from the seeded one, which is the level $\tfrac{1}{2}\overline{T}$ moving with
the query draw. Against that, $\Delta$ agrees in sign on all 5 scores. An unreplicated run reproduces the claim but does not pin down the
level.}
\label{tab:sweep}
\resizebox{\textwidth}{!}{%
\begin{tabular}{l ccccc c ccccc}
\toprule
& \multicolumn{5}{c}{\textit{CIFAR-100} ($C{=}50$)} && \multicolumn{5}{c}{\textit{ImageNet-1k} ($C{=}1000$)} \\
\cmidrule(lr){2-6}\cmidrule(lr){8-12}
$n_c$ & $25$ & $50$ & $100$ & $200$ & $400$ && $25$ & $50$ & $100$ & $200$ & $400$ \\
\midrule
Mean instability & 0.908 & 0.674 & 0.495 & 0.356 & 0.260 && 1.070 & 0.797 & 0.584 & 0.424 & 0.304 \\
\midrule
\texttt{knn\_std} & 0.4139 & 0.3112 & 0.2284 & 0.1649 & 0.1209 && 0.4728 & 0.3561 & 0.2638 & 0.1922 & 0.1387 \\
\texttt{lid} & 0.4142 & 0.3119 & 0.2299 & 0.1669 & 0.1227 && 0.4767 & 0.3580 & 0.2663 & 0.1953 & 0.1416 \\
\midrule
\textit{random} & \textit{0.4521} & \textit{0.3365} & \textit{0.2474} & \textit{0.1783} & \textit{0.1297} && \textit{0.5351} & \textit{0.3987} & \textit{0.2920} & \textit{0.2120} & \textit{0.1521} \\
\midrule
\texttt{energy} & 0.4801 & 0.3501 & 0.2567 & 0.1830 & 0.1324 && 0.5745 & 0.4201 & 0.3035 & 0.2202 & 0.1569 \\
\texttt{msp} & 0.4818 & 0.3512 & 0.2563 & 0.1832 & 0.1329 && 0.5732 & 0.4193 & 0.3018 & 0.2178 & 0.1545 \\
\texttt{maha} & 0.4762 & 0.3504 & 0.2574 & 0.1848 & 0.1348 && 0.5644 & 0.4163 & 0.3033 & 0.2196 & 0.1572 \\
\bottomrule
\end{tabular}}
\end{table*}

\section{The Two Targets Are Antipodal}
\label{app:orthogonal}

Section~\ref{sec:derma-flip} notes that the scores failing against verdict instability are optimized
against a different target. Table~\ref{tab:orthogonal} makes the point on CIFAR-100 and on ImageNet-1k. The left column of each
block is abstention against verdict instability and the right is abstention against
misclassification of the in-distribution classifier, the conventional selective-prediction target.
On CIFAR-100 MSP is the best score in the right column and the second worst in the left, and
\texttt{knn\_std} is the best in the left and the worst in the right. The two orders invert on
ImageNet-1k as well, and every score is beaten by a coin flip in the column it was not designed for.

\begin{table}[htbp]
\small
\centering
\caption{The two targets are close to antipodal. In each block the left column is abstention against
verdict instability and the right is abstention against misclassification of the in-distribution
classifier. Lower is better in both. Every score is beaten by a coin flip in the column it was not
designed for. Entries are means over the 10 seeds of Appendix~\ref{app:seeds}. The instability
columns are the AURC of Table~\ref{tab:worse} and the misclassification columns are measured on the
same queries at the same seeds.}
\label{tab:orthogonal}
\begin{tabular}{lrrcrr}
\toprule
& \multicolumn{2}{c}{\textit{CIFAR-100}} && \multicolumn{2}{c}{\textit{ImageNet-1k}} \\
\cmidrule(lr){2-3}\cmidrule(lr){5-6}
Score & AURC (instab.) & AURC (misclass.) && AURC (instab.) & AURC (misclass.) \\
\midrule
\texttt{knn\_std} & $\mathbf{0.1201}$ & $0.0756$ && $0.1376$ & $0.1780$ \\
\texttt{lid}      & $0.1220$ & $0.0744$ && $0.1406$ & $0.1810$ \\
\textit{random}   & \textit{0.1291} & \textit{0.0586} && \textit{0.1509} & \textit{0.1541} \\
\texttt{energy}   & $0.1317$ & $0.0270$ && $0.1554$ & $0.1027$ \\
\texttt{msp}      & $0.1321$ & $\mathbf{0.0193}$ && $0.1537$ & $0.0955$ \\
\texttt{maha}     & $0.1342$ & $0.0452$ && $0.1557$ & $0.1144$ \\
\bottomrule
\end{tabular}
\end{table}

\noindent
A deployment that abstains on MSP is buying accuracy
and selling stability, and one that abstains on \texttt{knn\_std} is doing the reverse. Neither is
wrong and neither is what the word suggests, and the choice is currently made by default.

\end{document}